\documentclass[10pt,twocolumn,letterpaper]{article}

\usepackage{iccv}
\usepackage{times}
\usepackage{epsfig}
\usepackage{graphicx}
\usepackage{amsmath}
\usepackage{amssymb}
\usepackage{paralist}

\iccvfinalcopy 

\input{taesup.sty}
\def\x{{\mathbf x}}

\newcommand{\Lb}{\mathbf{\Lambda}}
\newcommand{\Ellb}{\mathbf{L}}
\newcommand{\E}{\mathbb{E}}
\newcommand{\Z}{\mathbf{Z}}
\newcommand{\Xhb}{\hat{\mathbf{X}}}
\newcommand{\xh}{\hat{X}}
\newcommand{\N}{\mathbf{N}}

\newcommand{\naide}{{N-AIDE}~}

\newcommand{\naides}{{N-AIDE}$_{{S}}$~}

\newcommand{\fcaide}{\texttt{FC-AIDE}~}
\newcommand{\fcaideb}{\texttt{FC-AIDE}$_{\texttt{B}}$~}
\newcommand{\fcaides}{\texttt{FC-AIDE}$_{\texttt{S}}$~}
\newcommand{\fcaidesft}{\texttt{FC-AIDE}$_{\texttt{S+FT}}$~}
\newcommand{\fcaidebft}{\texttt{FC-AIDE}$_{\texttt{B+FT}}$~}

\usepackage{dsfont}
\usepackage{multirow}
\usepackage{wrapfig}
\usepackage{hyperref}
\hypersetup{
    colorlinks=true,
    linkcolor=red,
    filecolor=magenta,      
    urlcolor=cyan,
}

\begin{document}
\title{Fully Convolutional Pixel Adaptive Image Denoiser}

%

\author{
  Sungmin Cha\textsuperscript{\rm 1} and Taesup Moon\textsuperscript{\rm 1,2} \\
  \textsuperscript{\rm 1}Department of Electrical and Computer Engineering, \ \textsuperscript{\rm 2} Department of Artificial Intelligence\\
   Sungkyunkwan University, Suwon, Korea 16419\\
  \texttt{\{csm9493,tsmoon\}@skku.edu} \\
}

\maketitle
\ificcvfinal\thispagestyle{empty}\fi

\begin{abstract}
We propose a new image denoising algorithm, dubbed as Fully Convolutional Adaptive Image DEnoiser (FC-AIDE), that can learn from an offline supervised training set with a fully convolutional neural network as well as adaptively fine-tune the supervised model for each given noisy image. We significantly extend the framework of the recently proposed Neural AIDE, which formulates the denoiser to be context-based pixelwise mappings and utilizes the unbiased estimator of MSE for such denoisers. The two main contributions we make are; 1) implementing a novel fully convolutional architecture that boosts the base supervised model, and 2) introducing regularization methods for the adaptive fine-tuning such that a stronger and more robust adaptivity can be attained. 
As a result, FC-AIDE is shown to possess many desirable features; it outperforms the recent CNN-based state-of-the-art denoisers on all of the benchmark datasets we tested, and gets particularly strong for various challenging scenarios, e.g., with mismatched image/noise characteristics or with scarce supervised training data. The source code of our algorithm is available at  \href{https://github.com/csm9493/FC-AIDE-Keras}{\textcolor{magenta}{https://github.com/csm9493/FC-AIDE-Keras}}.




\end{abstract}


\section{Introduction}

Image denoising, which tries to recover a clean image from its noisy observation, is one of the oldest and most prevalent problems in image processing and low-level computer vision.  While numerous algorithms have been proposed over the past few decades, \eg, \cite{DonJoh95,Buadasetal05,bm3d,wnnm,ElaAha06,mai09}, the current throne-holders in terms of the average denoising performance are convolutional neural network (CNN)-based methods \cite{ZhaZuoCheMenZha17,MaoSheYan16,TaiYanLiuXu17}.

The main idea behind the CNN-based method is to treat denoising as a supervised regression problem; namely, first collect numerous clean-noisy image pairs as a supervised training set, then learn a regression function that maps the noisy image to the clean image. Such approach is relatively simple since it does not require any complex priors on the underlying clean images and let CNN figure out to learn the correct mapping from the vast training data. Despite the conceptual simplicity, several recent variations of the CNN-based denoisers using residual learning \cite{ZhaZuoCheMenZha17}, skip-connections \cite{MaoSheYan16}, and densely connected structure \cite{TaiYanLiuXu17} have achieved impressive state-of-the-art performances.


However, we stress that one apparent drawback exists in above methods; namely, they are solely based on offline batch training of CNN, hence, lack adaptivity to the given noisy image subject to denoising. Such absence of adaptivity, which is typically possessed in other prior or optimization-based methods, \eg, \cite{DonJoh95,Buadasetal05,bm3d,wnnm,ElaAha06,mai09}, can seriously deteriorate the denoising performance of the CNN-based methods in multiple practical scenarios in which various mismatches exist between the training data and the given noisy image. One category of such mismatch is the 
\emph{image  mismatch}, in which the image characteristics of the given noisy image are not well represented in the offline training set. The other category is \emph{noise mismatch}, in which the noise level or the distribution for the given noisy image is different from what the CNN has been trained for. 
Above drawbacks can be partially addressed by building a composite training set with multiple noise and image characteristics, \eg, the so-called blind training \cite{Lef18,ZhaZuoCheMenZha17}, but the limitation of such approach is evident since building large-scale supervised training set that sufficiently contains all the variations is not always possible in many applications.



To that end, we propose a new CNN-based denoising algorithm that can learn from an offline supervised training set, like other recent state-of-the-arts, as well as \emph{adaptively} fine-tune the denoiser for each given noisy image, like other prior or optimization based methods. 
The main vehicle for devising our algorithm, dubbed as FC-AIDE (Fully Convolutional Adaptive Image DEnoiser), is the framework recently proposed in \cite{ChaMoo18}. Namely, we formulate the denoiser to be context-based pixelwise mappings and obtain the SURE (Stein's Unbiased Risk Estimator)\cite{Ste81}-like estimated losses of the mean-squared errors (MSE) for the mappings. Then, the specific mapping for each pixel is learned with a neural network first by supervised training 
using the MSE,
then by adaptively fine-tuning the network with the given noisy image using the devised estimated loss.
While following this framework, we significantly improve the original Neural AIDE (N-AIDE) \cite{ChaMoo18} by making the following two main contributions. Firstly, we devise a novel fully \emph{convolutional} architecture instead of the simple fully-connected structure of N-AIDE so that the performance of the base supervised model can be boosted. 
Moreover, unlike \cite{ChaMoo18}, which only employed the pixelwise affine mappings, we also consider \emph{quadratic} mappings as well. 
Secondly, we introduce two regularization methods for the adaptive fine-tuning step: data augmentation and $\ell_2$-SP (Starting Point) regularization \cite{LiGraDav18}. While these techniques are well-known for general supervised learning setting, we utilize them such that the fine-tuned model does not \emph{overfit} to the estimated loss and generalize well to the MSE, leading to larger and more robust improvements of the fine-tuning step. 
With above two contributions, we show that FC-AIDE significantly surpasses the performance of the original N-AIDE as well as the CNN-based state-of-the-arts in several widely used benchmark datasets. Moreover, we highlight the effectiveness of the adaptivity of our algorithm in multiple challenging scenarios, e.g., with image/noise mismatches or with scarce supervised training data.

\section{Related Work} \label{sec:related}

There are vast number of literature on image denoising, among which the most relevant ones are elaborated here. \\
\noindent\textbf{Deep learning based denoisers} \ 
Following the pioneering work \cite{JaiSeu08,BurSchHar12,XieXuChe12}, recent deep learning based methods \cite{ZhaZuoCheMenZha17,MaoSheYan16,TaiYanLiuXu17} apply variations of CNN within the supervised regression framework and significantly surpassed the previous state-of-the-arts in terms of the denoising PSNR. More recently, \cite{Cruzetal18,LieWenFanLoyHua18} explicitly designed a non-local filtering process within the CNN framework and showed some more improvements with the cost of increased running time. 
For the cases with lack of clean data in the training data, \cite{LehMunHasLaiKarAitAil18} proposed a method that trains the CNN-based denoiser only with the noisy data, provided that two independent noise realizations for an image are available. \\
\noindent\textbf{Adaptive denoisers}\ \ In contrast to above CNN-based state-of-the-arts, which freeze the model parameters once the training is done, 
there are many approaches that can learn a denoising function adaptively from the given noisy image, despite some PSNR shortcomings.
Some of the classical methods are the filtering approach \cite{Buadasetal05,bm3d}, optimization-based methods with sparsity or low-rank assumptions \cite{ElaAha06,mai09,wnnm}, Wavelet transform-based \cite{DonJoh95}, and effective prior-based methods \cite{ZorWei11}. More recently, several work proposed to implement deep learning-based priors or regularizers, such as \cite{UlyVedLem18,YehLimChen18,LunzOktemSch18}, but their PSNR results still could not compete with the supervised trained CNN-based denoisers. \\
\indent Another branch of adaptive denoising is the universal denoising framework \cite{Dude}, in which no prior or probabilistic assumptions on the clean image are made, and only the noise is treated as random variables. 
The original \emph{discrete} denoising setting of \cite{Dude} was extended to grayscale image denoising in \cite{MotOrdRamSerWei11,Kamakshi}, but the performance was not very satisfactory.
A related approach is the SURE-based estimators that minimize the unbiased estimates of MSE, \eg, \cite{DonJoh95,Eld08}, but they typically select only a few tunable hyperparameters for the minimization. In contrast, using the unbiased estimate as an empirical risk in the empirical risk minimization (ERM) framework to learn an entire parametric model was first proposed in \cite{MooMinLeeYoo16,ChaMoo18} for denoising. To the best of the authors' knowledge, \cite{ChaMoo18} was the first deep learning-based method that has both the supervised learning capability and the adaptivity. 
\cite{SolChu18} also proposed to use SURE and Monte-Carlo approximation for
training a CNN-based denoiser, but the performance fell short of the supervised trained CNNs. \\
\noindent \textbf{Blind denoisers}\  
Most of above methods assume that the noise model and variance are known \emph{a priori}.
 To alleviate such strong assumption, the supervised, \emph{blindly} trained CNN models \cite{Lef18,ZhaZuoCheMenZha17} were proposed, and they show quite strong performance due to CNN's inherent robustness. But, those models also lack adaptivity, \ie, they cannot adapt to the specific noise realizations of the given noisy image.

\section{Problem Setting and Preliminaries}\label{sec:prelim}

\subsection{Notations and assumptions}
We generally follow \cite{ChaMoo18}, but introduce more compact notations. First, we denote $\mathbf{x}\in\mathbb{R}^n$ as the clean image and $\mathbf{Z}\in\mathbb{R}^n$ as its additive noise-corrupted version, namely, $\Z=\x+\mathbf{N}$. 
We 
make the following assumptions on $\mathbf{N}$;
\begin{compactitem}
	\item $\E(\N)=\mathbf{0}$ and $\text{Cov}(\N)=\sigma^2 \mathbf{I}_{n\times n}$.
	\item Each noise component, $N_i$, is independent and has a symmetric distribution.
\end{compactitem}
Note we do not necessarily assume the noise is identically distributed or Gaussian. Moreover, as in universal denoising, we treat the clean $\mathbf{x}$ as an individual image without any prior or probabilistic model and only treat $\mathbf{Z}$ as random, which is reflected in the upper case notation. 

A denoiser is generally denoted by $\hat{\mathbf{X}}(\Z)\in\mathbb{R}^n$, of which the $i$-th reconstruction, $\hat{X}_i(\Z)$, is a function of the entire noisy image $\Z$. 
 The standard loss function to measure the denoising quality is MSE, denoted by 
 \begin{align}
 \Lb_n(\x,\Xhb(\Z))\triangleq \frac{1}{n}\|\x-\Xhb(\Z)\|_2^2.\label{eq:mse}
 \end{align}


\subsection{N-AIDE and the unbiased estimator}\label{subsec:n-aide}

While $\hat{X}_i(\Z)$ can be any general function, we can consider a $d$-th order polynomial function form for the denoiser,
\begin{align}\hat{X}_i(\Z)=\sum_{m=0}^da_m(\Z^{-i})Z_i^m,\ \ \ i=1,\ldots,n,\label{eq:poly}\end{align}
in which $\Z^{-i}$ stands for the entire noisy image \emph{except} for $Z_i$, the $i$-th pixel, and $a_m(\Z^{-i})$ stands for the coefficient for the $m$-th order term of $Z_i$. Note \cite{ChaMoo18} focused on the case $d=1$, and in this paper, we consider the case $d=2$ as well. Note $\xh_i(\Z)$ can be a highly nonlinear function of $\Z$ since $a_m(\cdot)$'s can be any nonlinear functions of $\Z$.

 For (\ref{eq:poly}) with $d\in\{1,2\}$, by denoting $a_m(\Z^{-i})$ as $a_{m,i}$ for brevity, we can define an unbiased estimator of (\ref{eq:mse}) as
\begin{align}
&\Ellb_n(\Z,\Xhb(\Z);\sigma^2)\label{eq:est_loss}\\
\triangleq& \frac{1}{n}\|\Z-\Xhb(\Z)\|_2^2+\frac{\sigma^2}{n}\sum_{i=1}^n\Big[\sum_{m=1}^d2^ma_{m,i}Z_i^{m-1}-1\Big].\nonumber
\end{align}
Note 
(\ref{eq:est_loss}) does \emph{not} depend on $\x$, and the following lemma states the unbiasedness of (\ref{eq:est_loss}).
\begin{lemma}\label{lem1}
For any $\mathbf{N}$ with above assumptions and $\Xhb(\Z)$ that has the form (\ref{eq:poly}) with $d\in\{1,2\}$, 
\begin{eqnarray}
\E\Ellb_n(\Z,\Xhb(\Z);\sigma^2) = \E\Lb_n(\x,\Xhb(\Z)).\label{eq:unbiased}
\end{eqnarray}
Moreover, when $\mathbf{N}$ is white Gaussian, (\ref{eq:est_loss}) coincides with the SURE \cite{Ste81}.
\end{lemma}
\noindent\emph{Remark:} The proof of the lemma is given in the Supplementary Material, and it critically relies on the specific polynomial form of the denoiser in (\ref{eq:poly}). Namely, it exploits the fact that $\x$ is an individual image and $\{a_{m,i}\}$'s are conditionally independent of $Z_i$ given $\Z^{-i}$. Note (\ref{eq:unbiased}) holds for \emph{any} additive white noise with symmetric distribution, not necessarily only for Gaussian. (The symmetry condition is needed only for $d=2$ case.) Such property enables the strong adaptivity of our algorithm as shown below.  \\

N-AIDE in \cite{ChaMoo18} can be denoted by $\Xhb_{\texttt{N-AIDE}}(\wb,\Z)\in\mathbb{R}^n$, of which $\hat{X}_i(\wb,\Z)$ has the form (\ref{eq:poly}) with $d=1$ and 
\begin{eqnarray}a_{m,i}=a_m(\Z^{-i})\triangleq a_m(\wb,\Cb_{k\times k}^{-i})\label{eq:n-aide}
\end{eqnarray} for $m=0,1$. In (\ref{eq:n-aide}), $\Cb_{k\times k}^{-i}$ stands for the two-dimensional noisy $k\times k$ patch surrounding $Z_i$, but \emph{without} $Z_i$, and 
$\{a_m(\wb,\Cb_{k\times k}^{-i})\}_{m=0,1}$ are the outputs of a fully-connected neural network with parameter $\wb$ that takes $\Cb_{k\times k}^{-i}$ as input. Note $\wb$ does not depend on location $i$, hence, the denoising by N-AIDE is done in a sliding-window fashion; that is, the neural network subsequently takes $\Cb_{k\times k}^{- i}$ as an input and generates the affine mapping coefficients for $Z_i$ to obtain the reconstruction for each pixel $i$.

As mentioned in the Introduction, the training of the network parameter $\wb$ for N-AIDE was done in two stages. Firstly, for supervised training, a separate training set $\mathcal{D}=\{(\tilde{x}_i,\tilde{\Cb}_{i,k\times k})\}_{i=1}^N$ based on many clean-noisy image pairs, $(\tilde{\x},\tilde{\Z})$, is collected. Here, $\tilde{\Cb}_{i,k\times k}$ denote the full $k\times k$ patch \emph{including} the center pixel $\tilde{Z}_i$. Then, a supervised model, $\tilde{\wb}_{\text{sup}}$, the network parameters for a denoiser $\Xhb(\wb,\tilde{\Z})$ that has the form of (\ref{eq:n-aide}), is learned by minimizing the MSE on $\mathcal{D}$, \ie, $\Lb_N(\tilde{\x},\Xhb(\wb,\tilde{\Z}))$. Secondly, for a given noisy image $\Z$ subject to denoising, the adaptive fine-tuning is carried out by further minimizing the estimated loss $\Ellb_n(\Z,\Xhb(\wb,\Z);
\sigma^2)$ starting from $\tilde{\wb}_{\text{sup}}$. The fine-tuned model parameters, denoted by $\hat{\wb}_{\texttt{N-AIDE}}$, are then used to obtain the mapping for each pixel $i$ as in (\ref{eq:poly}) using (\ref{eq:n-aide}), and $\Z$ is pixelwise denoised with those affine mappings. Note maintaining the denoiser form (\ref{eq:poly}), and not the ordinary form of the direct mapping from the full noisy patch to the clean patch, is a key for enabling the fine-tuning.

\section{FC-AIDE}\label{sec:main}


\subsection{Fully convolutional QED architecture}\label{subsec:qed}

\begin{figure*}[h]
    \centering
    \subfigure[Overall architecture of FC-AIDE]{\label{fig:overall}
    \centering
    \includegraphics[width=0.7\linewidth]{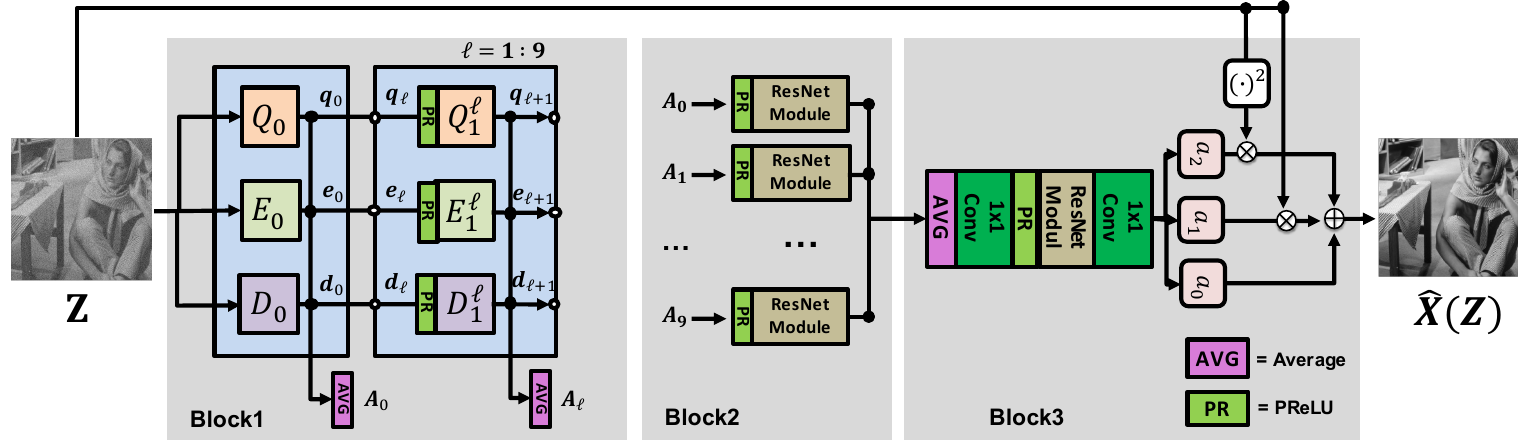}}
     \subfigure[ResNet Module]{\label{fig:resnet_module}
     \centering
    \includegraphics[width=0.2\linewidth]{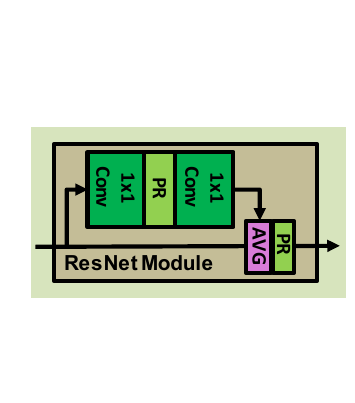}}
     \subfigure[Q-filters and the receptive field at layer 3]{\label{fig:q_filter}
    \includegraphics[width=0.32\linewidth]{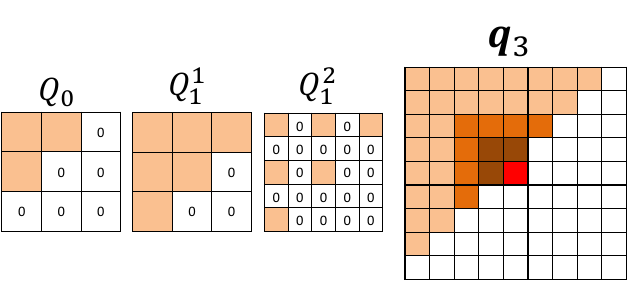}}
    \subfigure[E-filters and the receptive field at layer 3]{\label{fig:e_filter}
    \includegraphics[width=0.32\linewidth]{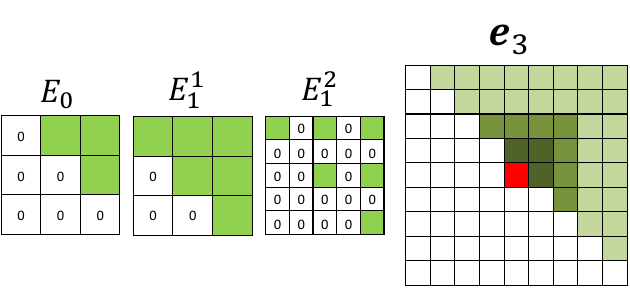}}
    \subfigure[D-filters and the receptive field at layer 3]{\label{fig:d_filter}
    \includegraphics[width=0.32\linewidth]{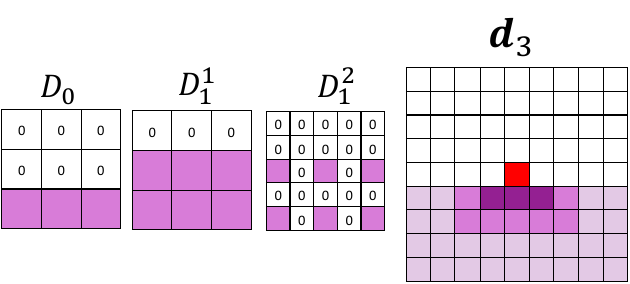}}
    \caption{Overall architecture of FC-AIDE and descriptions of the QED filter classes.}
    \label{fig:overall_architecture}\vspace{-.2in}
\end{figure*}


\noindent\textbf{Limitation of N-AIDE}\ \  While the performance of N-AIDE \cite{ChaMoo18} was encouraging, the final denoising performance was slightly worse (about $\sim$\ 0.15dB) than that of DnCNN-S \cite{ZhaZuoCheMenZha17}, which applies the ordinary supervised learning with CNN and residual learning. We believe such performance difference is primarily due to the simple fully-connected architecture used in \cite[Figure 1]{ChaMoo18} as opposed to the fully convolutional architectures in recent work, \eg, \cite{ZhaZuoCheMenZha17,MaoSheYan16,TaiYanLiuXu17}. 

In order to utilize the convolutional architecture for the N-AIDE framework, we first make an observation that the sliding-window nature of N-AIDE is indeed a convolution operation. Namely, once we use the masked $k\times k$ convolution filters with a \emph{hole} (\ie, the center set to 0) in the first layer and use the ordinary $1\times 1$ filters for the higher layers, the resulting fully convolutional network operating on a noisy image becomes equivalent to the fully-connected network of N-AIDE working in a sliding-window fashion. 
Note with the standard zero-padding, the output of the network has the same size as the input image and consists of two channels, one for each of $\{a_{m,i}\}_{m=0}^1$ for all location $i$.


From this observation, we identify a critical constraint for implementing a fully convolutional architecture for the denoisers that have the form (\ref{eq:poly});
 that is, the $i$-th pixel of \emph{any} feature maps at \emph{any} layer should \emph{not} depend on $Z_i$, and the filters at the output layer must keep the $1\times 1$ structure. Such property is necessary for maintaining the conditional independence between $\{a_{m,i}\}_{m=0}^d$ and $Z_i$ given $\Z^{-i}$, which is critical for the unbiasedness of $\Ellb_n(\Z,\Xhb(\Z);\sigma^2)$ shown in Lemma \ref{lem1}.
Due to this constraint, we can see that simply applying the vanilla convolutional architecture as in \cite{ZhaZuoCheMenZha17} is not feasible for extending N-AIDE beyond the $1\times 1$ structure.

\noindent\textbf{Overall architecture of FC-AIDE} \ To address above limitation of N-AIDE, we now propose FC-AIDE, denoted as $\Xhb_{\fcaide}(\wb,\Z)$, that has the fully convolutional architecture and gradually increases the receptive fields as the layer increases, while satisfying above mentioned constraint.
Moreover, we also extend to employ the \emph{quadratic} pixelwise mappings, \ie, use $d=2$ for (\ref{eq:poly}), to take into account of more nonlinearity in learning a denoiser. Namely, the $i$-th reconstruction of FC-AIDE becomes
\begin{align}
\hat{X}_{i,\fcaide}(\wb,\Z)=a_{2,i}Z_i^2+a_{1,i}Z_i+a_{0,i},\label{eq:fc_aide}
\end{align}
in which $\{a_{m,i}\}$'s are defined as (\ref{eq:n-aide}) for $m=0,1,2$.

Figure \ref{fig:overall} summarizes the overall architecture of our FC-AIDE. In Block 1, the three types of \emph{masked} dilated convolution filters, dubbed as Q-, E-, and D-filters, are stacked up to 10 layers, in which the specific filter forms for each class (up to layer 3) are depicted in Figure \ref{fig:q_filter}$\sim$\ref{fig:d_filter}, respectively. Note, as shown in Figure \ref{fig:overall}, the filters in each filter class is separately applied on its own input feature map, denoted as $\boldsymbol{q}_\ell$, $\boldsymbol{e}_\ell$, and $\boldsymbol{d}_\ell$ for $\ell=0,\ldots,9$. Due to the specific masked structure of the filters, the receptive field of each filter class for the $i$-th pixel in a feature map (colored in red in Figure \ref{fig:q_filter}$\sim$\ref{fig:d_filter}) gradually grows as the number of layers increases, while not containing $Z_i$. The example receptive fields at layer 3, i.e., $\boldsymbol{q}_3$, $\boldsymbol{e}_3$, and $\boldsymbol{d}_3$, for the red pixel are shown in Figure \ref{fig:q_filter}$\sim$\ref{fig:d_filter}. Now, since the receptive fields for the three filter classes cover complementary halfspaces of the input image, combining the three types of feature maps at layer $\ell$ (\eg, averaging) generates a feature map $A_\ell$, of which the $i$-th pixel depends on $\Cb_{k\times k}^{-i}$ with $k=3+\ell(\ell-1)$ for $\ell\geq1$. Note due to the \emph{dilated} filters for $\ell\geq 2$, the receptive fields grow quickly with the number of layers, while saving the number of parameters. 

Once the averaged feature maps $A_0,\ldots,A_9$ are generated from each layer, in Block 2 of Figure \ref{fig:overall}, the PReLU \cite{prelu} activation and ResNet Module, shown in Figure \ref{fig:resnet_module}, are applied to carry out additional nonlinear transformations of the feature maps. Finally, in Block 3, the feature maps from all layers (with different receptive fields) are averaged, 1x1 Conv with PReLU and one more ResNet Module is applied, and the three $1\times 1$ convolution filters are applied to generate the pixelwise coefficients, \ie, $\{a_{m,i}\}$'s for $m=0,1,2$ and for all $i$. Note our overall architecture is indeed fully convolutional, and since all the filters in Block 2 and 3 are $1\times 1$, the critical conditional independence constraint is satisfied.

We used 64 filters for all convolution layers in the model. The resulting number of parameters was about 820K, which is about 63\% smaller than that of N-AIDE in \cite{ChaMoo18}. However, thanks to the dilated filters, the receptive field size at the final layer was $93\times93$, much larger than N-AIDE. The training of FC-AIDE is done similarly as the description given in Section \ref{subsec:n-aide} for N-AIDE, except for the additional regularization methods for fine-tuning given in the next section.

\noindent \emph{Remark 1:} The conditional independence constraint similar to ours has also appeared in \cite{OorKalVinEspGraKav16}, in which ``one-sided'' contexts for sequentially generating images was considered. In contrast, FC-AIDE needs to  utilize the whole ``double-sided'' context, $\Cb_{k\times k}^{-i}$, for generating pixelwise mappings. 

\noindent \emph{Remark 2:} We note that filter classes other than our QED filters can also cover complementary halfspaces around $Z_i$, \eg, 4 filter classes that cover 4 perpendicular halfspaces surrounding a pixel. 
However, using three filter classes (as our QED filters or their $45^\circ$-rotated versions) turns out to be the most parameter-efficient choice.




\subsection{Regularization for adaptive fine-tuning}\label{subsec:fine-tune}



The fully-convolutional QED architecture in Section \ref{subsec:qed} expands the function approximation capability and improves the base supervised model benefitting from the abundant supervised training data. For the adaptive fine-tuning, however, the only ``training data'' is the given noisy image $\Z$, \ie, the test data, hence, is prone to \emph{overfitting} as the complexity of model increases. Namely, while the estimated loss (\ref{eq:est_loss}), based on $\mathbf{Z}$, is minimized during fine-tuning, the true MSE (\ref{eq:mse}), based on both $\mathbf{x}$ and $\mathbf{Z}$, may not be minimized as much. Note the notion of overfitting and generalization for the fine-tuning is different from the ordinary one; \ie, while the ordinary supervised learning cares about the performance with respect to the \emph{unseen} test data, our fine-tuning cares about the performance with respect to the \emph{unseen} clean data. 
In order to address this overfitting issue, we implement two regularization schemes for adaptive fine-tuning: data augmentation and $\ell_2$-SP regularization \cite{LiGraDav18}.

First, for data augmentation, 
 we consider $\mathcal{A}(\Z)$, which is the augmented dataset that consists of $\Z$ and its horizotally, vertically, and both horizontally and vertically flipped versions. 
Then, we define $\Ellb_n^{\text{aug}}(\cdot)$ as the average of the estimated losses on $\mathcal{A}(\Z)$, \ie, 
\be
\Ellb_n^{\text{aug}}(\Z,\wb;\sigma^2)\triangleq \frac{1}{4}\sum_{\Z^{(j)}\in\mathcal{A}(\Z)}
\Ellb_n\big(\Z^{(j)},\hat{\Xb}(\wb,\Z^{(j)});\sigma^2\big).\nonumber
\ee
Then, the $\ell_2$-SP regularization adds the squared $\ell_2$-norm penalty on the deviation from the supervised trained FC-AIDE model, $\tilde{\wb}_{\text{sup}}$, and modify the objective function for fine-tuning as
\be
\Ellb_n^{\text{aug}}(\Z,\wb;\sigma^2)+\lambda\|\wb-\tilde{\wb}_{\text{sup}}\|_2^2,\label{eq:augment}
\ee
in which $\lambda$ is a trade-off hyperparameter. This simple additional penalty, also considered in \cite{LiGraDav18} in the context of transfer learning, can be interpreted as imposing a \emph{prior}, which is learned from the supervised training set, on the network parameters; note the similarity of (\ref{eq:augment}) to the formulation of other prior-based denoising methods.

With above two regularization methods, we fine-tune the supervised model $\tilde{\wb}_{\text{sup}}$ by minimizing (\ref{eq:augment}) and obtain the final weight parameters $\hat{\wb}_{\fcaide}$. The denoising is then carried out by averaging the results obtained from applying $\hat{\wb}_{\fcaide}$ in (\ref{eq:fc_aide}) to the 4 images in $\mathcal{A}(\Z)$ separately. Note our data augmentation is different from the ordinary one in supervised learning, since we are \emph{training} with the test data before testing. 
In Section \ref{subsec:ablation}, we analyze the effects\
 of both regularization techniques on fine-tuning more in details.

\section{Experimental results}\label{sec:experiments}

\subsection{Data and experimental setup}\label{subsec:data}




\noindent\textbf{Training details}\ \ For the supervised training, we exactly followed \cite{ZhaZuoCheMenZha17} and used 400 publicly available natural images of size $180\times 180$ for building training dataset. We randomly sampled total 20,500 patches of size $120\times120$ from the images. Note the amount of information in our training data, in terms of the number of pixels, is roughly the same as that of DnCNN-S in \cite{ZhaZuoCheMenZha17} (204,800 patches of size $40\times40$), which uses 10 times more patches that are 9 times smaller than ours. Moreover, we used standard Gaussian noise augmentation for generating every mini-batch of clean-noisy patch pair for training of both $\sigma$-specific and the blindly trained models. 
Learning rates of 0.001 and 0.0003 were used for supervised training and fine-tuning, respectively, and Adam \cite{KinBa15} optimizer was used. Learning rate decay was used only for the supervised training, and dropout or BatchNorm were not used. All experiments used Keras 2.2.0 with Tensorflow 1.8.0 and NVIDIA GTX1080TI.



\noindent\textbf{Evaluation data}\ \ We first used five benchmark datasets, Set5, Set12\cite{ZhaZuoCheMenZha17}, BSD68\cite{foe} Urban100\cite{Huang2015}, and Manga109\cite{manga109}, to objectively compare the performance of FC-AIDE with other state-of-the-arts for Gaussian denoising. 
Among the benchmarks, Set12 and BSD68 contain general natural grayscale images of which characteristics are similar to that of the training set. In contrast, Set5 (visualized in the Supplementary Material), Urban100 (images with many self-similar patterns), and Manga109 (cartoon images) contain images that are quite different from the training data. 
We tested with five different noise levels, $\sigma=\{15,25,30,50,75\}$. Overall, we used the standard metrics, PSNR(dB) and SSIM, for evaluation.


  In addition, we generated two additional datasets to evaluate and compare the adaptivities of the algorithms. 
Firstly, \emph{Medical/Gaussian} is a set of 50 medical images (collected from the CT, MRI, and X-ray modalities) of size $400\times400$, corrupted by Gaussian noise with $\sigma=\{30,50\}$. The images were obtained from the open repositories \cite{xray,CT,mri}. Clearly, their characteristics are radically different from natural images. Secondly, \emph{BSD68/Laplacian} is generated by corrupting BSD68 by Laplacian noise with $\sigma=\{30,50\}$.
These datasets are built to test the adaptivity for the image and noise mismatches, respectively, since all the comparing CNN-based methods, including FC-AIDE, are supervised trained only on natural images with Gaussian noise. 

\begin{table*}[t]\caption{PSNR(dB)/SSIM on benchmarks with Gaussian noise. The best and the second best are denoted in red and blue, respectively.}\label{table:benchmark}
\centering
\smallskip\noindent
\resizebox{\linewidth}{!}{
    \begin{tabular}{|c|c|c|c|c|c|c|c||c|c|c|c|}

\hline
Data     & Noise                                                                                                       & BM3D                                                                                                             & RED                                                                             & DnCNN$_S$                                                                                                        & DnCNN$_B$                                                                                                        & Memnet                                                                          & N-AIDE$_{S+FT}$                                                                                                   & \fcaides                                                                                                      & \fcaidesft                                                                                                           & \texttt{FC-AIDE$_{\texttt{B}}$}                                                                                                      & \texttt{FC-AIDE$_{\texttt{B+FT}}$}                                                                                    
 \\   
\hline
\hline
Set5    & 
\begin{tabular}[c]{@{}c@{}}$\sigma$=15\\ $\sigma$=25\\ $\sigma$=30\\ $\sigma$=50\\ $\sigma$=75\end{tabular} & 
\begin{tabular}[c]{@{}c@{}}29.64/0.8983\\ 26.47/0.8983\\ 25.32/0.8764\\ 23.20/0.8139\\ 21.21/0.7409\end{tabular} & 
\begin{tabular}[c]{@{}c@{}}-\\ -\\ 25.14/0.8953\\ 23.02/0.8161\\ -\end{tabular} & 
\begin{tabular}[c]{@{}c@{}}30.22/0.9479\\ 27.01/0.9072\\ 25.95/0.8882\\ 22.67/0.7969\\ 20.47/0.6899\end{tabular} & 
\begin{tabular}[c]{@{}c@{}}28.76/0.9364\\ 26.14/0.8948\\ 25.15/0.8735\\ 22.58/0.7949\\ 17.30/0.5437\end{tabular} & 
\begin{tabular}[c]{@{}c@{}}-\\ -\\ 25.97/0.8909\\ 23.17/0.8205\\ -\end{tabular} & 
\begin{tabular}[c]{@{}c@{}}30.23/0.9480\\ 27.05/0.9083\\ 26.01/0.8896\\ 23.08/0.8160\\ 20.95/0.7323\end{tabular} & 
\begin{tabular}[c]{@{}c@{}}29.96/0.9461\\ 26.91/0.9083\\ 25.85/0.8890\\ 22.37/0.7992\\ 20.38/0.7091\end{tabular} & 
\begin{tabular}[c]{@{}c@{}}\textbf{\textcolor{red}{30.78/0.9518}}\\ \textbf{\textcolor{red}{27.88/0.9191}}\\ \textbf{\textcolor{red}{26.86/0.9038}}\\ \textbf{\textcolor{blue}{24.20}}/\textbf{\textcolor{red}{0.8482}}\\ \textbf{\textcolor{blue}{22.20}}/\textbf{\textcolor{red}{0.7889}}\end{tabular}  & 
\begin{tabular}[c]{@{}c@{}}29.67/0.9453\\ 26.75/0.9070\\ 25.72/0.8882\\ 22.93/0.8117\\ 20.62/0.7176\end{tabular} & 
\begin{tabular}[c]{@{}c@{}}\textbf{\textcolor{blue}{30.69/0.9514}}\\ \textbf{\textcolor{blue}{27.83/0.9182}}\\ \textbf{\textcolor{blue}{26.80/0.9030}}\\ \textbf{\textcolor{red}{24.23}}/\textbf{\textcolor{blue}{0.8475}}\\ \textbf{\textcolor{red}{22.22}}/\textbf{\textcolor{blue}{0.7857}}\end{tabular}
\\\hline
\hline
Set12    & 
\begin{tabular}[c]{@{}c@{}}$\sigma$=15\\ $\sigma$=25\\ $\sigma$=30\\ $\sigma$=50\\ $\sigma$=75\end{tabular} & 
\begin{tabular}[c]{@{}c@{}}32.15/0.8856\\ 29.67/0.8327\\ 28.74/0.8085\\ 26.55/0.7423\\ 24.68/0.6670\end{tabular} & 
\begin{tabular}[c]{@{}c@{}}-\\ -\\ 29.68/0.8378\\ 27.32/0.7748\\ -\end{tabular} & 
\begin{tabular}[c]{@{}c@{}}32.83/0.8964\\ 30.40/0.8513\\ 29.53/0.8321\\ 27.16/0.7667\\ 25.27/0.7001\end{tabular} & 
\begin{tabular}[c]{@{}c@{}}32.50/0.8899\\ 30.15/0.8435\\ 29.30/0.8233\\ 26.94/0.7528\\ 17.64/0.2802\end{tabular} & 
\begin{tabular}[c]{@{}c@{}}-\\ -\\ 29.62/\textbf{\textcolor{red}{0.8374}}\\ \textbf{\textcolor{blue}{27.36}}/\textbf{\textcolor{red}{0.7791}}\\ -\end{tabular} & 
\begin{tabular}[c]{@{}c@{}}32.58/0.8918\\ 30.12/0.8420\\ 29.24/0.8214\\ 26.84/0.7492\\ 24.90/0.6749\end{tabular} & 
\begin{tabular}[c]{@{}c@{}}32.71/0.8962\\ 30.32/0.8521\\ 29.47/0.8334\\ 27.16/0.7698\\ 25.37/0.7094\end{tabular} & 
\begin{tabular}[c]{@{}c@{}}\textbf{\textcolor{red}{32.99/0.9006}}\\ \textbf{\textcolor{red}{30.57/0.8557}}\\ \textbf{\textcolor{red}{29.74}}/\textbf{\textcolor{blue}{0.8373}}\\ \textbf{\textcolor{red}{27.42}}/\textbf{\textcolor{blue}{0.7768}}\\ \textbf{\textcolor{red}{25.61/0.7170}}\end{tabular}  & 
\begin{tabular}[c]{@{}c@{}}32.48/0.8929\\ 30.16/0.8490\\ 29.33/0.8303\\ 27.04/0.7636\\ 24.93/0.6703\end{tabular} & 
\begin{tabular}[c]{@{}c@{}}\textbf{\textcolor{blue}{32.91/0.8995}}\\ \textbf{\textcolor{blue}{30.51/0.8545}}\\ \textbf{\textcolor{blue}{29.63}}/0.8353\\ 27.29/0.7693\\ \textbf{\textcolor{blue}{25.39/0.6980}}\end{tabular} \\ \hline
\hline
BSD68    & 
\begin{tabular}[c]{@{}c@{}}$\sigma$=15\\ $\sigma$=25\\ $\sigma$=30\\ $\sigma$=50\\ $\sigma$=75\end{tabular} & 
\begin{tabular}[c]{@{}c@{}}31.07/0.8717\\ 28.56/0.8013\\ 27.74/0.7727\\ 25.60/0.6866\\ 24.19/0.6216\end{tabular} & 
\begin{tabular}[c]{@{}c@{}}-\\ -\\ \textbf{\textcolor{blue}{28.45}}/0.7987\\ 26.29/0.7124\\ -\end{tabular} & 
\begin{tabular}[c]{@{}c@{}}31.69/0.8869\\ 29.19/0.8202\\ 28.36/0.7925\\ 26.19/0.7027\\ 24.64/0.6240\end{tabular} & 
\begin{tabular}[c]{@{}c@{}}31.40/0.8804\\ 28.99/0.8132\\ 28.17/0.7847\\ 26.05/0.6934\\ 17.91/0.2856\end{tabular} & 
\begin{tabular}[c]{@{}c@{}}-\\ -\\ 28.42/0.7915\\  \textbf{\textcolor{blue}{26.34}}/ \textbf{\textcolor{red}{0.7190}}\\ -\end{tabular} & 
\begin{tabular}[c]{@{}c@{}}31.49/0.8825\\ 28.99/0.8137\\ 28.15/0.7842\\ 25.98/0.6911\\ 24.40/0.6101\end{tabular} & 
\begin{tabular}[c]{@{}c@{}}31.67/0.8885\\ 29.20/0.8246\\ 28.38/0.7974\\ 26.27/0.7127\\ 24.77/0.6402\end{tabular} & 
\begin{tabular}[c]{@{}c@{}}\textbf{\textcolor{red}{31.78/0.8907}}\\ \textbf{\textcolor{red}{29.31/0.8281}}\\ \textbf{\textcolor{red}{28.49/0.8014}}\\ \textbf{\textcolor{red}{26.38}}/\textbf{\textcolor{blue}{0.7181}}\\ \textbf{\textcolor{red}{24.89/0.6477}}\end{tabular}  &
\begin{tabular}[c]{@{}c@{}}31.53/0.8859\\ 29.11/0.8213\\ 28.30/0.7937\\ 26.18/0.7063\\ 24.41/0.6151\end{tabular} & 
\begin{tabular}[c]{@{}c@{}}\textbf{\textcolor{blue}{31.71/0.8897}}\\ \textbf{\textcolor{blue}{29.26/0.8267}}\\ 28.44/\textbf{\textcolor{blue}{0.7995}}\\ 26.32/0.7132\\ \textbf{\textcolor{blue}{24.75/0.6340}}\end{tabular} \\ \hline
\hline
Urban100 & 
\begin{tabular}[c]{@{}c@{}}$\sigma$=15\\ $\sigma$=25\\ $\sigma$=30\\ $\sigma$=50\\ $\sigma$=75\end{tabular} & 
\begin{tabular}[c]{@{}c@{}}31.61/0.9301\\ 28.76/0.8773\\ 27.71/0.8520\\ 25.22/0.7686\\ 23.20/0.6804\end{tabular} & 
\begin{tabular}[c]{@{}c@{}}-\\ -\\ 28.58/0.8743\\ 25.83/0.7946\\ -\end{tabular} & 
\begin{tabular}[c]{@{}c@{}}32.32/0.9375\\ 29.41/0.8884\\ 28.38/0.8659\\ 25.66/0.7843\\ 23.57/0.6973\end{tabular} & 
\begin{tabular}[c]{@{}c@{}}31.75/0.9257\\ 29.04/0.8787\\ 28.07/0.8556\\ 25.42/0.7732\\ 18.35/0.4254\end{tabular} & 
\begin{tabular}[c]{@{}c@{}}-\\ -\\ 28.57/0.8720\\ 26.00/0.8021\\ -\end{tabular} & 
\begin{tabular}[c]{@{}c@{}}31.96/0.9340\\ 29.11/0.8861\\ 28.10/0.8622\\ 25.40/0.7767\\ 23.31/0.6843\end{tabular} & 
\begin{tabular}[c]{@{}c@{}}31.98/0.9311\\ 29.15/0.8871\\ 28.16/0.8646\\ 25.55/0.7885\\ 23.61/0.7099\end{tabular} & 
\textbf{\textcolor{red}{\begin{tabular}[c]{@{}c@{}}32.85/0.9448\\ 30.05/0.9053\\ 29.06/0.8867\\ 26.42/0.8178\\ 24.42/0.7469\end{tabular}}} & 
\begin{tabular}[c]{@{}c@{}}31.54/0.9299\\ 28.91/0.8850\\ 27.99/0.8635\\ 25.46/0.7845\\ 23.28/0.6902\end{tabular} & 
\textbf{\textcolor{blue}{\begin{tabular}[c]{@{}c@{}}32.65/0.9433\\ 29.91/0.9033\\ 28.91/0.8837\\ 26.23/0.8087\\ 24.11/0.7248\end{tabular}}} \\ \hline
\hline
Manga109 & 
\begin{tabular}[c]{@{}c@{}}$\sigma$=15\\ $\sigma$=25\\ $\sigma$=30\\ $\sigma$=50\\ $\sigma$=75\end{tabular} & 
\begin{tabular}[c]{@{}c@{}}32.02/0.9362\\ 29.00/0.8917\\ 27.91/0.8691\\ 25.24/0.7943\\ 23.20/0.7143\end{tabular} &
\begin{tabular}[c]{@{}c@{}}-\\ -\\ 29.52/0.9011\\ 26.50/0.8358\\ -\end{tabular} &
\begin{tabular}[c]{@{}c@{}}33.52/0.9489\\ 30.40/0.9121\\ 29.33/0.8945\\ 26.38/0.8237\\ 24.04/0.7394\end{tabular} & 
\begin{tabular}[c]{@{}c@{}}33.01/0.9406\\ 30.15/0.9041\\ 29.09/0.8851\\ 26.17/0.8146\\ 18.82/0.4308\end{tabular} &
\begin{tabular}[c]{@{}c@{}}-\\ -\\ 29.55/0.9003\\ 26.64/0.8403\\ -\end{tabular} &
\begin{tabular}[c]{@{}c@{}}33.07/0.9450\\ 30.04/0.9071\\ 28.96/0.8880\\ 26.02/0.8160\\ 23.75/0.7355\end{tabular} & 
\begin{tabular}[c]{@{}c@{}}33.24/0.9457\\ 30.24/0.9112\\ 29.18/0.8935\\ 26.28/0.8281\\ 24.06/0.7577\end{tabular} &
\textbf{\textcolor{red}{\begin{tabular}[c]{@{}c@{}}33.86/0.9524\\ 30.80/0.9193\\ 29.75/0.9041\\ 26.79/0.8428\\ 24.54/0.7789\end{tabular}}}  &
\begin{tabular}[c]{@{}c@{}}32.83/0.9433\\ 30.02/0.9086\\ 29.02/0.8913\\ 26.19/0.8246\\ 23.76/0.7281\end{tabular} &
\textbf{\textcolor{blue}{\begin{tabular}[c]{@{}c@{}}33.70/0.9521\\ 30.72/0.9191\\ 29.65/0.9030\\ 26.66/0.8361\\ 24.33/0.7588\end{tabular}}} \\ \hline

\multicolumn{2}{|l|}{Training data ratio}  & -            & \textbf{x4.23}        & x1 & x1           & \textbf{x4.23}        & x1             & x1            & x1  & x1            & x1       \\
\hline
\end{tabular}}
\vspace{-.2in}
\end{table*}

\noindent \textbf{Comparing methods}\ \ The baselines we used were: BM3D \cite{bm3d}, RED \cite{MaoSheYan16}, Memnet \cite{TaiYanLiuXu17}, DnCNN-S and DnCNN-B \cite{ZhaZuoCheMenZha17}, and N-AIDE \cite{ChaMoo18}. For DnCNN and N-AIDE, we used the available source codes to reproduce \emph{both} training (on the same supervised training data as FC-AIDE) and denoising, and for BM3D/RED/MemNet, we downloaded the models from the authors' website and carried out the denoising in our evaluation datasets. Thus, all the numbers in our tables are \emph{fairly} comparable. 


We denote \fcaidesft as our final model obtained after the supervised training and adaptive fine-tuning. For comparison purpose, we also report the results of the \fcaide with subscripts \texttt{S}, \texttt{B}, and \texttt{B+FT}, which stand for the supervised-only model, the blindly trained supervised model, and the blindly trained model fine-tuned with true $\sigma$, respectively. For N-AIDE, we also report the \texttt{S+FT} scheme, but the fine-tuning was done without any regularization methods.
For the blind supervised models, DnCNN-B and \fcaideb were all trained with Gaussian noise with $\sigma\in[0,55]$.
The stopping epochs for all our \fcaide models (both supervised and fine-tuned) as well as $\lambda$ in (\ref{eq:augment}) were selected from a separate validation set that is composed of 32 images from BSD \cite{berkeley}. All details regarding the experimental settings are given in the Supplementary Material.

\subsection{Denoising results on the benchmark datasets}

Table \ref{table:benchmark} shows the results on the 5 benchmark datasets, supervised training data size ratios, and the average denoising time per image for all comparing models. 
 For RED and Memnet, we only have results for $\sigma=30,50$, since the models for other $\sigma$ were not available. 
There are several observations that we can make. 
Firstly, we note \fcaidesft outperforms all the comparing state-of-the-arts on most datasets in terms of PSNR/SSIM. Moreover, the gains of \fcaidesft against the strongest baselines get significantly larger for the datasets that have the \emph{image mismatches}, \ie, Set5, Urban100, and Manga109. 
This is primarily due to the effectiveness of the adaptive fine-tuning step that significantly improves \fcaides. The additional results that highlight such adaptivity are also given in Section \ref{subsec:adaptivity} and \ref{subsec:ablation}. 
 Secondly, we note that \fcaidesft uses much less supervised training data than RED and Memnet in terms of the number of pixels. Again, a more detailed analysis on the data efficiency of \fcaidesft is given in Section \ref{subsec:adaptivity}. 
 Thirdly, it is clear that \fcaidesft is much better than N-AIDE$_{S+FT}$, which has the fully-connected structure and no regularizations for fine-tuning, confirming our contributions given in Section \ref{subsec:qed} and \ref{subsec:fine-tune}. \vspace{-.1in}
 \begin{figure}[H]
\centering
\includegraphics[width=0.9\linewidth]{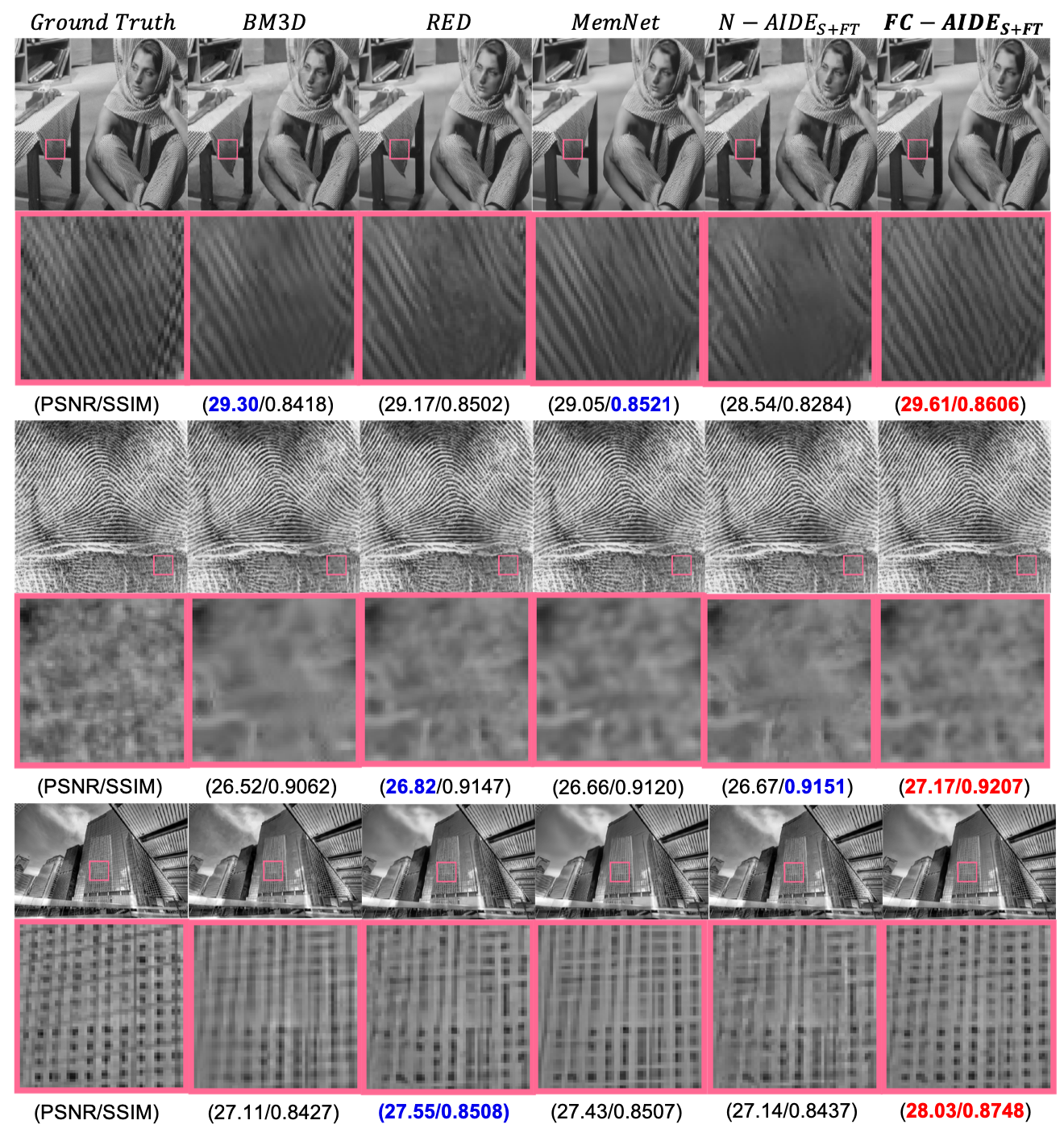}
\caption{Denoising results ($\sigma=30$) for \emph{Barbara}, \emph{F.print}, and \emph{Image60} in Urban100.}
\label{fig:vis}\vspace{-.1in}
\end{figure}
\noindent Fourthly, we note \fcaidebft also is quite strong, \ie, gets very close to \fcaidesft and mostly outperforms the other baselines with \emph{matched} noise for supervised models. This result is interesting since it suggests maintaining just a single blindly trained supervised model, \fcaideb, is sufficient as long as the true $\sigma$ is available for the fine-tuning. Note also DnCNN-B fails dramatically for $\sigma=75$, which is outside the noise levels that DnCNN-B is trained for, but \fcaidebft corrects most of such \emph{noise mismatch}. 
 Finally, the denoising time of \fcaidesft is larger than that of \texttt{FC-AIDE}$_{\texttt{S}}$, which is a cost to pay for the adaptivity. 

 Figure \ref{fig:vis} visualizes the denoising results for $\sigma=30$, particularly for images with many self-similar patterns. A notable example is \emph{Barbara}, in which other CNN baselines are worse than BM3D, but \fcaidesft significantly outperforms all others. Overall, we see that \fcaidesft performs very well on the images with self-similarities \emph{without} any explicit non-local operations as in \cite{LieWenFanLoyHua18,Cruzetal18}.

\subsection{Effects of adaptive fine-tuning}\label{subsec:adaptivity}


Here, we give additional results that highlight the three main scenarios in which the adaptivity of \fcaidesft gets particularly effective.

\begin{figure}[h]
    \centering
    \includegraphics[width=.7\linewidth]{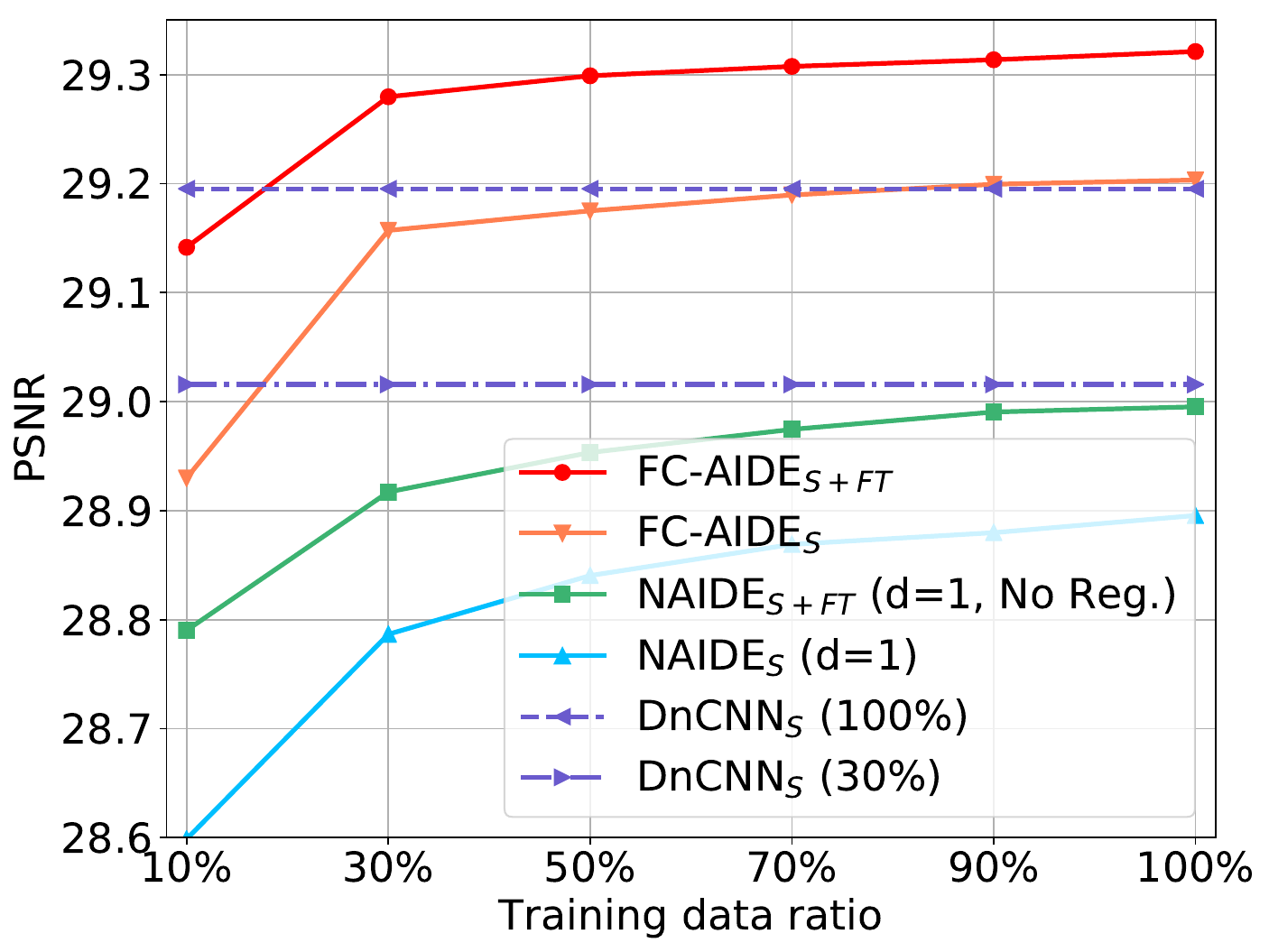}
    \caption{Data efficiency over DnCNN\cite{ZhaZuoCheMenZha17} for BSD68 ($\sigma=25$).}
    \label{fig:data-efficient}\vspace{-.1in}
\end{figure}

\noindent\textbf{Data scarcity} \ Figure \ref{fig:data-efficient} compares the PSNR of \fcaidesft with DnCNN-S on BSD68 ($\sigma=25$) with varying training data size; \ie, the horizontal axis represents the relative training data size compared to that used for training DnCNN-S in \cite{ZhaZuoCheMenZha17}, in terms of the number of pixels. The performance of N-AIDE$_S$ and N-AIDE$_{S+FT}$ are also shown for comparison purpose. From the figure, we observe that \fcaidesft surpasses DnCNN-S (100\%) with using only 30\% of the training data due to the two facts; the base supervised model \fcaides is more data-efficient (\ie, \fcaides with 30\% data outperforms DnCNN-S (30\%)), and the fine-tuning gives another 0.1dB PSNR boost. 

\noindent\textbf{Image mismatch}\ \ Table \ref{table:medical} gives denoising results on \emph{Medical/Gaussian}, which consists of 50 medical images with radically different characteristics compared to the natural images in the supervised training set of RED, Memnet, and \fcaides. From the table, we again see that \fcaidesft outperforms RED and Memnet despite \fcaides being slightly worse than them, thanks to the adaptivity of fine-tuning. Moreover, we consider a scenario in which there is only a small number of \emph{matched} supervised training data available; \ie, \texttt{FC-AIDE$_{\texttt{S(M)}}$} in Table \ref{table:medical} is a supervised model trained with only 10 medical images. In this case, we observe that \texttt{FC-AIDE$_{\texttt{S(M)}}$} is even worse than above three \emph{mismatched} supervised models, due to the small training data size. However, via fine-tuning, we observe \texttt{FC-AIDE$_{\texttt{S(M)+FT}}$} surpasses all other models and achieves the best PSNR, which shows the effectiveness of the adaptivity for fixing the image mismatches. 

\vspace{-.05in}
\begin{table}[h]\caption{PSNR(dB)/SSIM on \emph{Medical/Gaussian}. Color as before.}\label{table:medical50}

\centering
\smallskip\noindent
\resizebox{\linewidth}{!}{
    \begin{tabular}{|c|c|c|c||c|c|c|c|}
\hline
Noise                                                                                                                         & RED                                                                                                                           & Memnet                                                              & N-AIDE$_{S+FT}$                                                     & \fcaides                                                       & \fcaidesft                                                    & \texttt{FC-AIDE$_{\texttt{S(M)}}$}                                                       & \texttt{FC-AIDE$_{\texttt{S(M)+FT}}$}                                                    \\ \hline
\hline
\begin{tabular}[c]{@{}c@{}}$\sigma=30$\\ $\sigma=50$\end{tabular} &
\begin{tabular}[c]{@{}c@{}}35.12/0.9005\\ 32.78/0.8660\end{tabular} & 
\begin{tabular}[c]{@{}c@{}}35.02/0.8986\\ 32.87/0.8691\end{tabular} &
\begin{tabular}[c]{@{}l@{}}34.70/0.8920\\ 32.23/0.8499\end{tabular} &
\begin{tabular}[c]{@{}l@{}}35.01/0.8980\\ 32.74/0.8641\end{tabular} &
\textbf{\textcolor{blue}{\begin{tabular}[c]{@{}l@{}}35.26/0.9030\\ 32.99/0.8703\end{tabular}}} & 
\begin{tabular}[c]{@{}l@{}}34.96/0.8993\\ 32.56/0.8628\end{tabular} &
\textbf{\textcolor{red}{\begin{tabular}[c]{@{}l@{}}35.37/0.9050\\ 33.06/0.8727\end{tabular}}} \\ \hline
\end{tabular}}
    \vspace{-.15in}
    \label{table:medical}
\end{table}



\begin{table*}[h]\caption{PSNR(dB)/SSIM on BSD68 with Laplacian noise. The best and the second best are denoted in red and blue, respectively.}\label{table:laplacian}

\centering
\smallskip\noindent
\resizebox{\linewidth}{!}{
    \begin{tabular}{|c|c|c|c|c|c|c||c|c|c|c|c|c|}
\hline
Noise                                                                                                       & BM3D                                                                                                             & RED                                                                             & DnCNN-S                                                                                                        & DnCNN-B                                                                                                        & Memnet                                                                           & N-AIDE$_{S+FT}$                                                                                                   & \texttt{FC-AIDE$_{\texttt{S}}$}                                                                                                      & \texttt{FC-AIDE$_{\texttt{S+FT}}$}                                                                                                          & \texttt{FC-AIDE$_{\texttt{B}}$}                                                                                                      & \texttt{FC-AIDE$_{\texttt{B+FT}}$}                                                                                                 & \texttt{FC-AIDE$_{\texttt{S(L)}}$}                                                                                                   & \texttt{FC-AIDE$_{\texttt{S(L)+FT}}$}                                                                                                
\\ \hline
\hline
\begin{tabular}[c]{@{}c@{}}$\sigma$=30\\ $\sigma$=50\end{tabular} & 
\begin{tabular}[c]{@{}c@{}}27.48/0.7564\\ 25.52/0.6669\end{tabular} & 
\begin{tabular}[c]{@{}c@{}}28.18/0.7887\\ 26.10/0.7030\end{tabular} & 
\begin{tabular}[c]{@{}c@{}}28.01/0.7783\\ 25.90/0.6878\end{tabular} &
\begin{tabular}[c]{@{}c@{}}27.69/0.7650\\ 25.68/0.6769\end{tabular} &
\begin{tabular}[c]{@{}c@{}}28.26/0.7916\\ 26.13/0.7098\end{tabular} &
\begin{tabular}[c]{@{}c@{}}28.08/0.7797\\ 25.91/0.6858\end{tabular} &
\begin{tabular}[c]{@{}c@{}}28.28/0.7917\\ 26.17/0.7067\end{tabular} & 
\textbf{\textcolor{red}{\begin{tabular}[c]{@{}c@{}}28.42/0.7983\\ 26.31/0.7123\end{tabular}}} & 
\begin{tabular}[c]{@{}c@{}}28.12/0.7886\\ 25.92/0.6954\end{tabular} & 
\textbf{\textcolor{blue}{\begin{tabular}[c]{@{}c@{}}28.41/0.7979\\ 26.27/0.7077\end{tabular}}} & 
\begin{tabular}[c]{@{}c@{}}28.63/0.8076\\ 26.64/0.7293\end{tabular} & 
\textbf{\begin{tabular}[c]{@{}c@{}}28.70/0.8090\\ 26.70/0.7317\end{tabular}} \\ \hline
\end{tabular}}\vspace{-.2in}
\end{table*}

\begin{figure}[h]
\centering
\includegraphics[width=.9\linewidth]{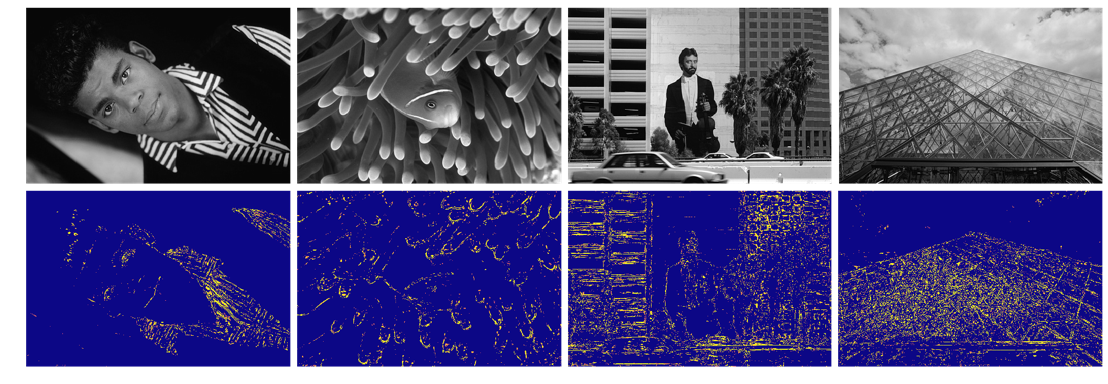}
\caption{Row 1: Top 4 images in BSD68 ($\sigma=25$) that had the most PSNR improvements by \fcaidesft over \fcaides Row 2: 
The pixels that had most improvements. 
}
\label{fig:bsd-nonlocal}\vspace{-.1in}
\end{figure}

From Table \ref{table:benchmark} and Figure \ref{fig:data-efficient}, one may think the gain of \fcaidesft over \fcaides is relatively small for BSD68 (\eg, $0.1$dB on average for $\sigma=25$). 
We stress, however, that this is due to the similarity between the training data and BSD68. 
That is, Figure \ref{fig:bsd-nonlocal} shows the top 4 images in BSD68 ($\sigma=25$) that \fcaidesft had the most improvement over \texttt{FC-AIDE}$_{\texttt{S}}$; the improvement for each image was 0.53dB, 0.49dB, 0.38dB, and 0.29dB, respectively, which are much higher than the average. The pixels that had the most MSE improvements are shown as yellow pixels in the second row. We clearly observe that these images are with many self-similar patterns and
 see that \fcaidesft gets particularly strong primarily at pixels with those patterns. Images with specific self-similar patterns can be considered as another form of \emph{image mismatch}, and we confirm the effectiveness of our fine-tuning.


\noindent\textbf{Noise mismatch} \ \ Table \ref{table:laplacian} shows the denoising results on \emph{BSD68/Laplacian}. Since all the supervised models in the table are trained with Gaussian noise, the setting corresponds to the \emph{noise mismatch} case. As ideal upper bounds, we also report the performance of \texttt{FC-AIDE$_{\texttt{S(L)}}$} and \texttt{FC-AIDE$_{\texttt{S(L)+FT}}$}, which stand for the supervised model trained with Laplacian noise-corrupted data and its fine-tuned model, respectively. We can see that among models using the mismatched supervised models, \fcaidesft again achieves the best denoising performance followed by \texttt{FC-AIDE$_{\texttt{B+FT}}$}, without any information on the noise distribution other that $\sigma$. Moreover, we observe the PSNR gap between \fcaides and \texttt{FC-AIDE$_{\texttt{S(L)}}$} are much reduced after fine-tuning, and the gaps between \fcaidesft and the other baselines widened compared to those in Table \ref{table:benchmark}.

\subsection{Ablation study and analyses}\label{subsec:ablation}

Here, we give more detailed analyses justifying our modeling choices given in Section \ref{subsec:qed} and \ref{subsec:fine-tune}.

\begin{figure}[h]
    \centering
    \subfigure[Model architecture variations]{\label{fig:ablation-arch}
     \centering
    \includegraphics[width=.48\linewidth]{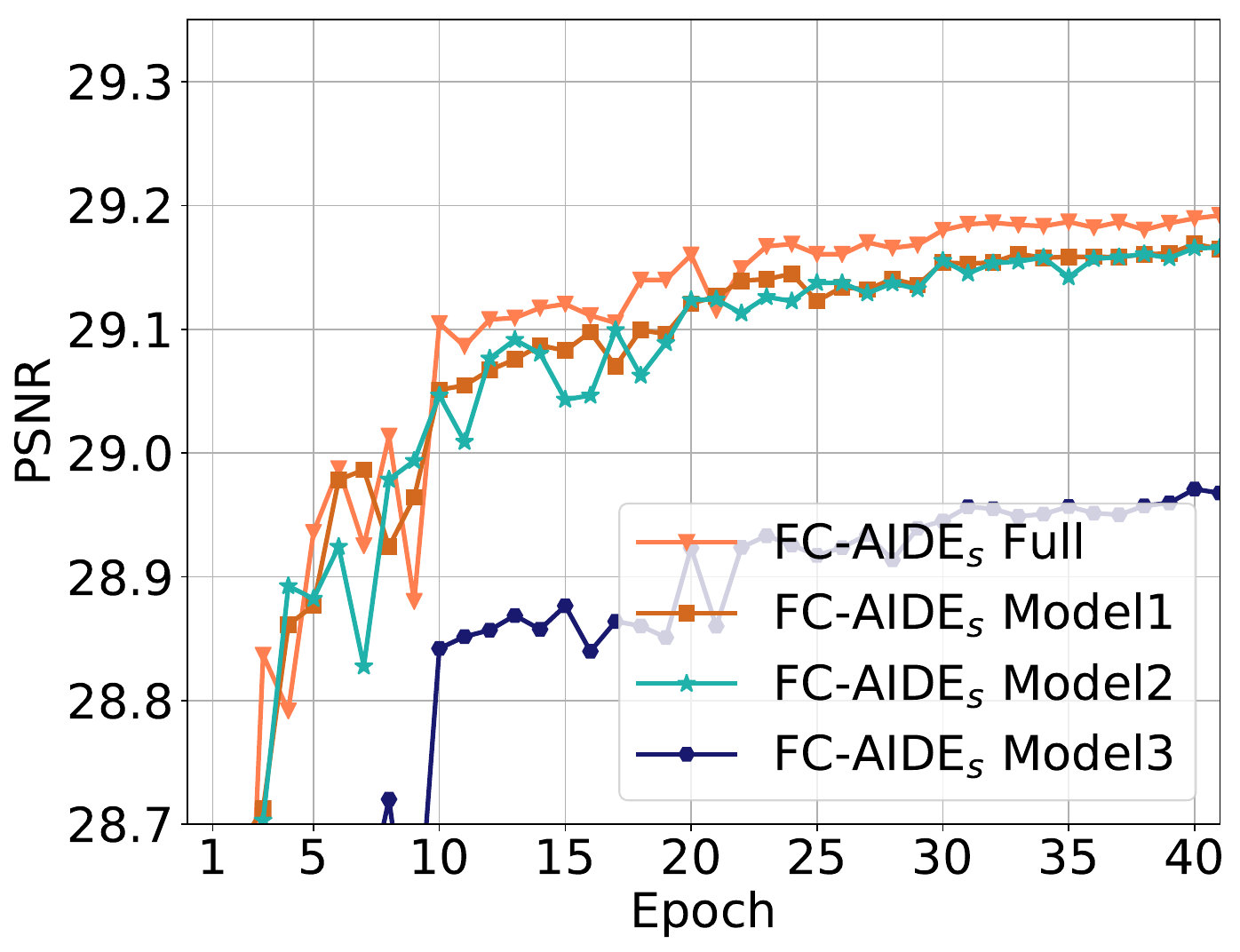}}
    \subfigure[Improvements over \naide \cite{ChaMoo18}.]{\label{fig:ablation_naide}
    \centering
    \includegraphics[width=.48\linewidth]{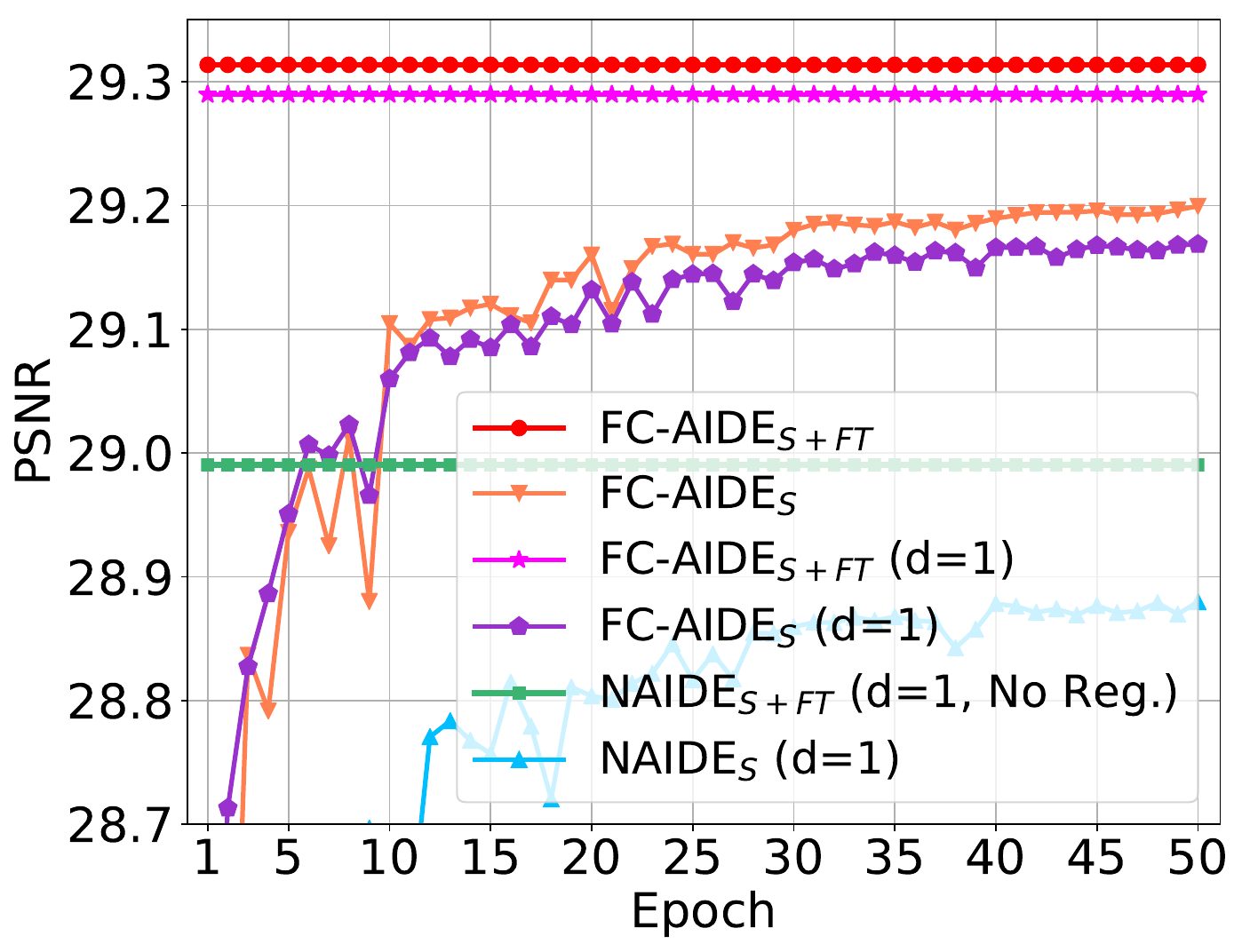}}
    \vspace{-.05in}
    \caption{Ablation studies for \fcaide on BSD68 ($\sigma=25$). }
    \label{fig:ablation-studies}\vspace{-.1in}
\end{figure}

\noindent\textbf{Ablation study on model architecture} \ \ Figure \ref{fig:ablation-studies} shows several ablation studies on BSD68 ($\sigma=25$) with varying model architectures.
\vspace{-.1in}
Firstly, Figure \ref{fig:ablation-arch} shows the PSNR of \fcaides with several varying architectures with respect to the training epochs. ``Full'' stands for the model architecture in Figure \ref{fig:overall}, ``Model 1'' is without the ResNet Modules in Block 2, ``Model 2'' is without the ResNet Module in Block 3, and ``Model 3'' is without both ResNet Modules. The figure shows the ResNet Modules in Figure \ref{fig:overall} are all critical in our model. Secondly, Figure \ref{fig:ablation_naide} compares \fcaidesft to its $d=1$ version as well as to \naide \cite{ChaMoo18}. We clearly observe the benefit of our QED architecture over the fully-connected architecture, since the supervised-only \fcaides ($d=1$) significantly outperforms both \naides and N-AIDE$_{S+FT}$. Moreover, we observe the quadratic mappings for \fcaidesft also are beneficial in further improving the PSNR. 

\vspace{-.05in}
\begin{table}[h]\caption{PSNR(dB) on BSD68 ($\sigma=25$).}
\centering
\smallskip\noindent
\resizebox{\linewidth}{!}{
    \begin{tabular}{|c||c|c||c|c||c|}
\hline
Data$\backslash$Alg.     & \fcaides & \fcaidesft   &  \texttt{No-QED}$_\texttt{S}$ & \texttt{No-QED}$_\texttt{S+FT}$& \texttt{FC-AIDE}$_{\texttt{FT}}$ \\ \hline \hline
BSD68 &  29.20 & \textbf{29.31} & 29.16& 21.93 & 23.76\\ \hline
\end{tabular}}
    \label{table:no-qed}\vspace{-.1in}
\end{table}

In Table \ref{table:no-qed}, we justify our QED architecture, which satisfies the conditional independence constraint mentioned in Section \ref{subsec:qed}, by comparing with a CNN that learns the same polynomial mapping (\ref{eq:poly}) with vanilla convolution filters, which violates the constraint. Such model, dubbed as \texttt{No-QED}, had the same number of layers and receptive field as those of \texttt{FC-AIDE}. We observe that while the supervised model, \texttt{NO-QED}$_{\texttt{S}}$ gets quite close to \texttt{FC-AIDE}$_{\texttt{S}}$, the fine-tuned model \texttt{NO-QED}$_{\texttt{S+FT}}$ dramatically deteriorates. This shows the conditional independence constraint is indispensable for the fine-tuning and justifies our QED architecture. Moreover, Table \ref{table:no-qed} also shows the performance of \texttt{FC-AIDE}$_{\texttt{FT}}$, which only carries out fine-tuning with randomly initialized parameters. The result clearly shows the importance of supervised training for FC-AIDE.

\begin{figure}[h]
    \centering
      \subfigure[Data augmentation]{\label{fig:aug_ft}
     \centering
    \includegraphics[width=.47\linewidth]{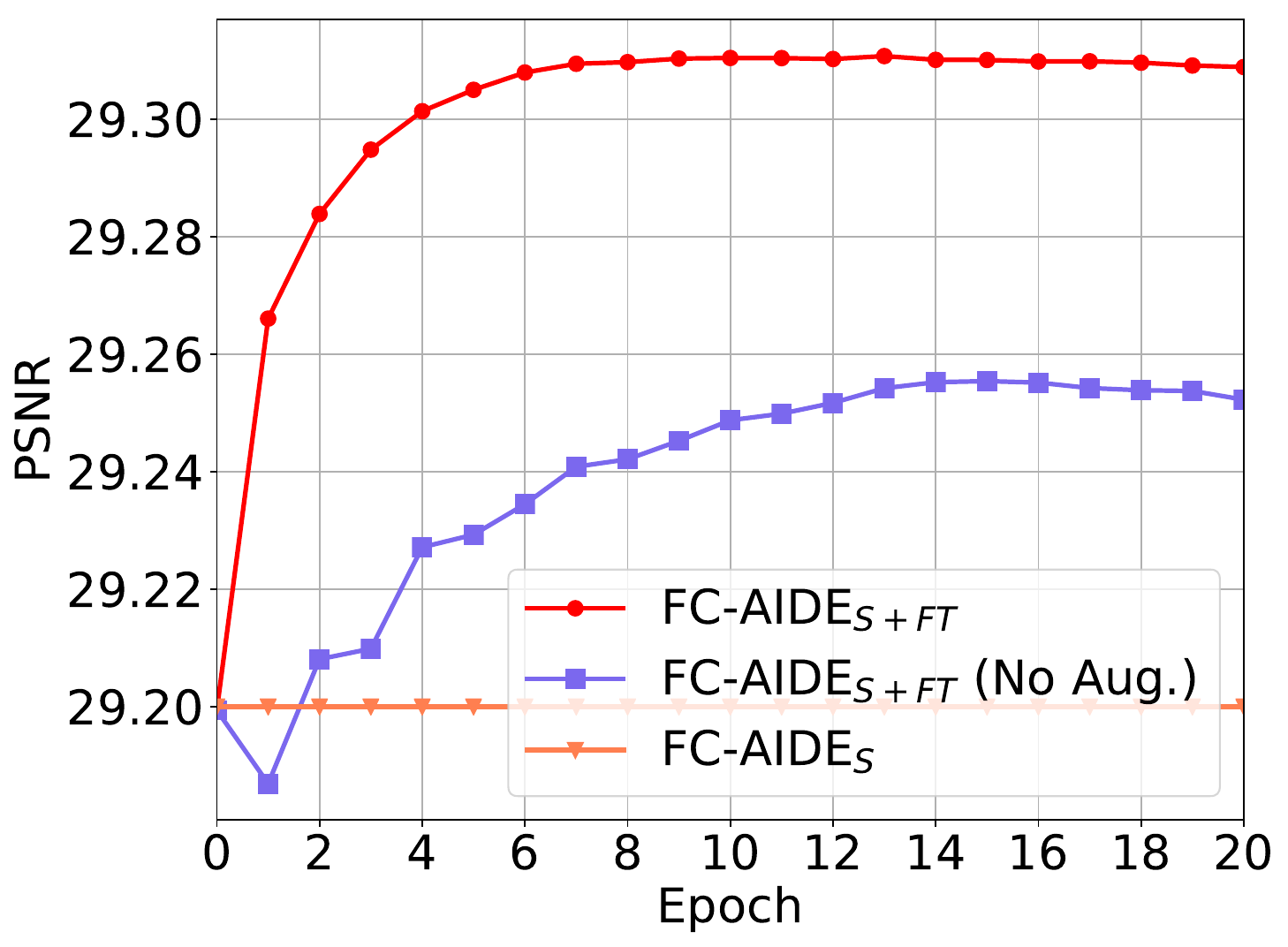}}
    \subfigure[$\ell_2$-SP]{\label{fig:l2reg}
    \centering
    \includegraphics[width=.49\linewidth]{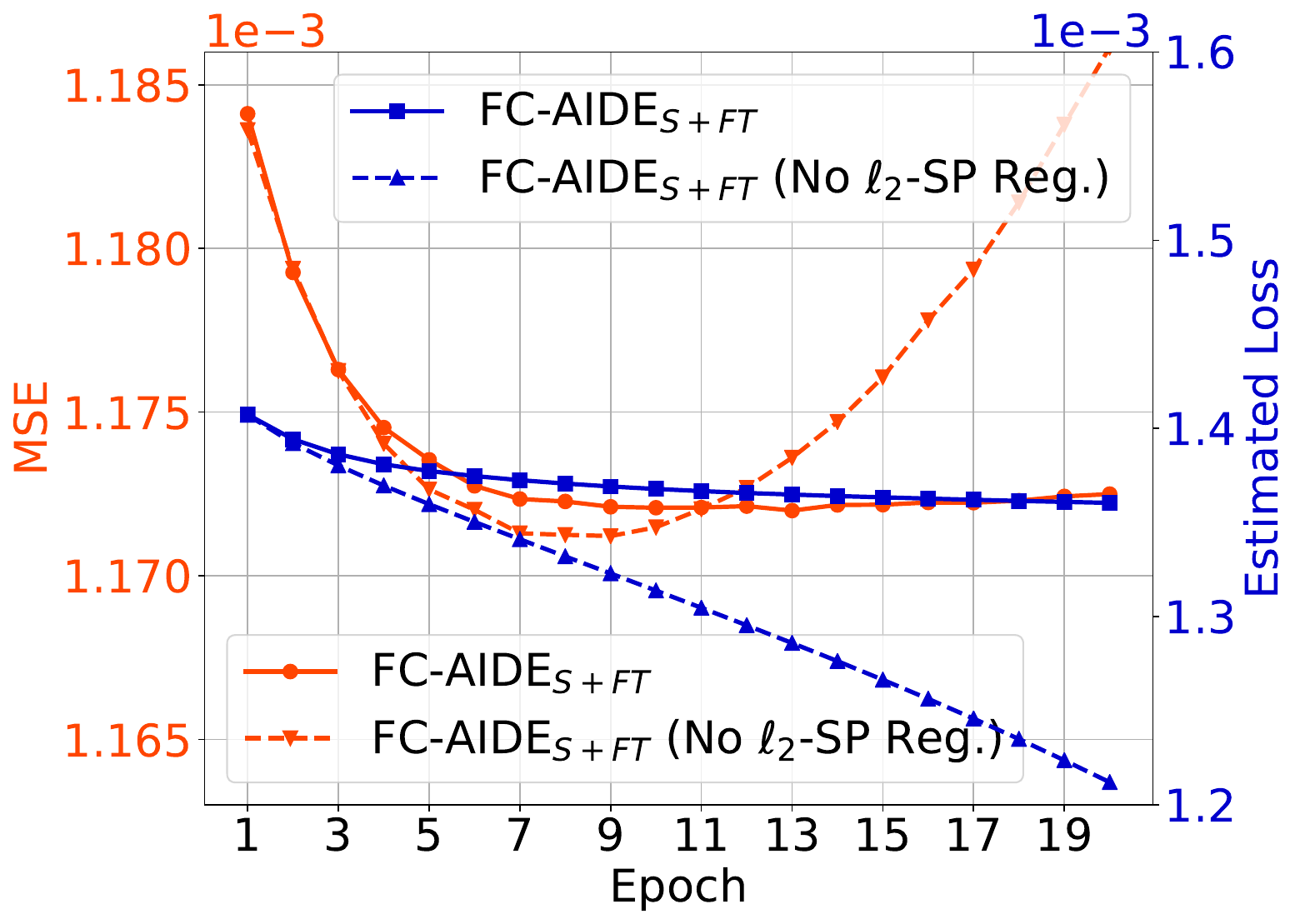}}
    \vspace{-.05in}
    \caption{Effects of regularization methods for BSD68 ($\sigma=25$).}
    \label{fig:reg_aug}\vspace{-.1in}
\end{figure}

\noindent\textbf{Effect of regularization}
Figure \ref{fig:reg_aug} shows the effect of each regularization method in Section \ref{subsec:fine-tune} on BSD68 ($\sigma=25$). Firstly, Figure \ref{fig:aug_ft} compares the PSNR of \fcaidesft with and without the data augmentation. We clearly observe that the data augmentation gives a further boost of PSNR compared to just using single $\Z$. Secondly, Figure \ref{fig:l2reg} shows MSE (orange) and estimated loss (blue) during fine-tuning with and without $\ell_2$-SP. We can clearly observe that when there is no $\ell_2$-SP, the trends of MSE and the estimated loss diverges (\ie, \emph{overfitting} occurs), but with $\ell_2$-SP, minimizing the estimated loss generalizes well to minimizing the MSE with robustness.


 
\vspace{-.1in}



\begin{figure}[h]
    \centering
    \includegraphics[width=.95\linewidth]{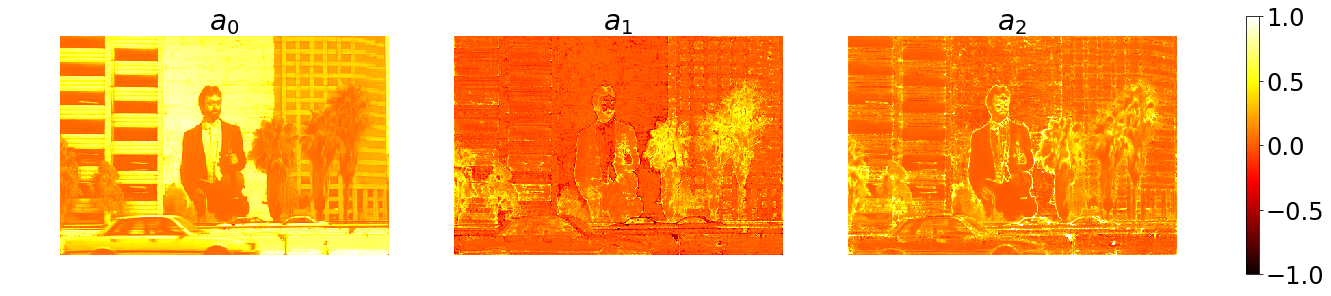}
    \caption{Visualization of $a_0$, $a_1$ and $a_2$ for \emph{Image13} in BSD68. }
    \label{fig:a1a2a3_error}\vspace{-.2in}
\end{figure}

\noindent\textbf{Visualization of polynomial mapping} \ \ Figure \ref{fig:a1a2a3_error} visualizes the pixelwise polynomial coefficients $\{a_{m,i}\}_{m=0}^2$ learned for \emph{Image13} of BSD68 ($\sigma=25$). We note the coefficient values for the higher order terms are relatively small compared to $\{a_{0,i}\}$'s.
 But, we observe they become more salient particularly for the high frequency parts of the image, \ie, the edges. The effects of the polynomial coefficients on denoising is given in the Supplementary Material. \vspace{-.1in}

\section{Conclusion}\label{sec:conclusion} 
We proposed FC-AIDE that can \emph{both} supervised-train and adaptively fine-tune the CNN-based pixelwise denoising mappings. While surpassing the strong recent baselines, we showed that the adaptivity of our method can resolve various mismatch as well as data scarce scenarios commonly encountered in practice. Possible future research directions include extending our method to other general image restoration problems, \eg, image super-resolution, beyond denoising and devise a full blind denoising method that can also estimate the noise $\sigma$.

\section*{Acknowledgement}\label{sec:Acknowledgement} 

\noindent This work is supported in part by the ICT R\&D Program [2016-0-00563], AI Graduate School Support Program [2019-0-00421], and ITRC Support Program [2019-2018-0-01798] of MSIT / IITP of the Korean government.



\newpage

\bibliographystyle{ieee_fullname}
\bibliography{bibfile}

\begin{thebibliography}{10}\itemsep=-1pt

\bibitem{CT}
National cancer institute clinical proteomic tumor analysis consortium (cptac).
\newblock https://doi.org/10.7937/k9/tcia.2018.oblamn27.

\bibitem{Buadasetal05}
A. Buades, B. Coll, and J.~M. Morel.
\newblock A review of image denoising algorithms, with a new one.
\newblock {\em SIAM Journal on Multiscale Modeling and Simulation: A SIAM
  Interdisciplinary Journal}, 2005.

\bibitem{BurSchHar12}
H. Burger, C. Schuler, and S. Harmeling.
\newblock Image denoising: {C}an plain neural networks compete with {BM3D}?
\newblock In {\em Computer Vision and Pattern Recognition (CVPR)}, 2012.

\bibitem{ChaMoo18}
Sungmin Cha and Taesup Moon.
\newblock Neural adaptive image denoiser.
\newblock In {\em IEEE ICASSP}, 2018.

\bibitem{mri}
Jun Cheng.
\newblock Brain tumor dataset, 2017.

\bibitem{Cruzetal18}
Cristovao Cruz, Alessandro Foi, Vladimir Katkovnik, and Karen Egiazarian.
\newblock Nonlocality-reinforced convolutional neural networks for image
  denoising.
\newblock {\em https://arxiv.org/pdf/1803.02112.pdf}, 2018.

\bibitem{bm3d}
K. Dabov, A. Foi, V. Katkovnik, and K. Egiazarian.
\newblock Image denoising by sparse 3-d transform-domain collaborative
  filtering.
\newblock {\em IEEE Trans. Image Processing}, 16(8):2080--2095, 2007.

\bibitem{DonJoh95}
D. Donoho and I. Johnstone.
\newblock Adapting to unknown smoothness via wavelet shrinkage.
\newblock {\em Journal of American Statistical Association},
  90(432):1200--1224, 1995.

\bibitem{ElaAha06}
M. Elad and M. Aharon.
\newblock Image denoising via sparse and redundant representations over learned
  dictionaries.
\newblock {\em IEEE Trans. Image Processing}, 54(12):3736--3745, 2006.

\bibitem{Eld08}
Y. Eldar.
\newblock Rethinking biased estimation: {I}mproving maximum likelihood and the
  {C}ramer-{R}ao bound.
\newblock {\em Foundations and Trends in Signal Processing}, 1(4):305--449,
  2008.

\bibitem{wnnm}
S. Gu, L. Zhang, W. Zuo, and X. Feng.
\newblock Weighted nuclear norm minimization with applicaitons to image
  denoising.
\newblock In {\em Computer Vision and Pattern Recognition (CVPR)}, 2014.

\bibitem{prelu}
Kaiming He, Xiangyu Zhang, Shaoqing Ren, and Jian Sun.
\newblock Delving deep into rectifiers: Surpassing human-level performance on
  imagenet classification.
\newblock In {\em CVPR}, 2015.

\bibitem{Huang2015}
J-B Huang, A. Singh, and N. Ahuja.
\newblock Single image super-resolution from transformed self-exemplars.
\newblock In {\em CVPR}, 2015.

\bibitem{JaiSeu08}
V. Jain and H.S. Seung.
\newblock Natural image denoising with convolutional networks.
\newblock In {\em NIPS}, 2008.

\bibitem{KinBa15}
D. Kingma and J. Ba.
\newblock Adam: A method for stochastic optimization.
\newblock In {\em International Conference on Learning Representations (ICLR)},
  2015.

\bibitem{Lef18}
Stamatios Lefkimmiatis.
\newblock Universal denoising networks: {A} novel {CNN} architecture for image
  denoising.
\newblock In {\em CVPR}, 2018.

\bibitem{LehMunHasLaiKarAitAil18}
Jaakko Lehtinen, Jacob Munkberg, Jon Hasselgren, Samuli Laine, Tero Karras,
  Miika Aittala, and Timo Aila.
\newblock Noise2{N}oise: {L}earning image restoration without clean data.
\newblock In {\em ICML}, 2018.

\bibitem{LiGraDav18}
X. Li, Y. Grandvalet, and F. Davoine.
\newblock Explicit inductive bias for transfer learning with convolutional
  networks.
\newblock In {\em ICML}, 2018.

\bibitem{LieWenFanLoyHua18}
Ding Liu, Bihan Wen, Yuchen Fan, Chen~C. Loy, and Thomas~S. Huang.
\newblock Non-local recurrent network for image restoration.
\newblock In {\em NIPS}, 2018.

\bibitem{LunzOktemSch18}
S. Lunz, O. {\"O}ktem, and C.-B. Sch{\"o}nlieb.
\newblock Adversarial regularizers in inverse problems.
\newblock In {\em NIPS}, 2018.

\bibitem{mai09}
J. Mairal, F. Bach, J. Ponce, G. Sapiro, and A. Zisserman.
\newblock Non-local sparse models for image restoration.
\newblock In {\em International Conference on Computer Vision (ICCV)}, 2009.

\bibitem{MaoSheYan16}
X. Mao, C. Shen, and Y-B. Yang.
\newblock Image restoration using very deep convolutional encoder-decoder
  networks with symmetric skip connections.
\newblock {\em Neural Information Processing Systems (NIPS)}, 2016.

\bibitem{berkeley}
D. Martin, C. Fowlkes, D. Tal, and J. Malik.
\newblock A database of human segmented natural images and its application to
  evaluating segmentation algorithms and measuring ecological statistics.
\newblock In {\em International Conference on Computer Vision (ICCV)}, 2001.

\bibitem{manga109}
Y. Matsui, K. Ito, Y. Aramaki, A. Fujimoto, T. Ogawa, T. Yamasaki, and K.
  Aizawa.
\newblock Sketch-based manga retrieval using {M}anga109 dataset.
\newblock {\em Multimed. Tools Appl.}, 76(21811), 2017.

\bibitem{MooMinLeeYoo16}
T. Moon, S. Min, B. Lee, and S. Yoon.
\newblock Neural universal discrete denosier.
\newblock In {\em Neural Information Processing Systems (NIPS)}, 2016.

\bibitem{MotOrdRamSerWei11}
Giovanni Motta, Erik Ordentlich, Ignacio Ramirez, Gadiel Seroussi, and
  Marcelo~J. Weinberger.
\newblock The i{DUDE} framework for grayscale image denoising.
\newblock {\em IEEE Trans. Image Processing}, 20:1--21, 2011.

\bibitem{foe}
S. Roth and M.J Black.
\newblock Field of experts.
\newblock {\em International Journal of Computer Vision}, 82(2):205--229, 2009.

\bibitem{Kamakshi}
K. Sivaramakrishnan and T. Weissman.
\newblock Universal denoising of discrete-time continuous-amplitude signals.
\newblock {\em {IEEE} Trans. Inform. Theory}, 54(12):5632--5660, 2008.

\bibitem{SolChu18}
Shakarim Soltanayev and Se~Young Chun.
\newblock Training deep learning based denoisers without ground truth data.
\newblock In {\em NIPS}, 2018.

\bibitem{Ste81}
C. Stein.
\newblock Estimation of the mean of a multivariate normal distribution.
\newblock {\em The Annals of Statistics}, 9(6):1135--1151, 1981.

\bibitem{TaiYanLiuXu17}
Ying Tai, Jian Yang, Xiaoming Liu, and Chunyan Xu.
\newblock Memnet: {A} persistent memory network for image restoration.
\newblock In {\em ICCV}, 2017.

\bibitem{UlyVedLem18}
Dmitry Ulyanov, Andrea Vedaldi, and Victor Lempitsky.
\newblock Deep image prior.
\newblock In {\em CVPR}, 2018.

\bibitem{OorKalVinEspGraKav16}
A{\"a}ron van~den Oord, Nal Kalchbrenner, Oriol Vinyals, Lasse Esleholt, Alex
  Graves, and Koray Kavukcuoglu.
\newblock Conditional image generation with {P}ixel{CNN} decoders.
\newblock In {\em NIPS}, 2016.

\bibitem{xray}
Xiaosong Wang, Yifan Peng, Le Lu, Zhiyong Lu, MohammadhadiBagheri, and
  Ronald~M. Summers.
\newblock Chestx-ray8: Hospital-scale chest x-ray database and benchmarks on
  weakly-supervised classification and localization of common thorax diseases.
\newblock In {\em CVPR}, 2017.

\bibitem{Dude}
T. Weissman, E. Ordentlich, G. Seroussi, S. Verdu, and M. Weinberger.
\newblock Universal discrete denoising: {K}nown channel.
\newblock {\em {IEEE} Trans. Inform. Theory}, 51(1):5--28, 2005.

\bibitem{XieXuChe12}
J. Xie, L. Xu, and E. Chen.
\newblock Image denoising and inpainting with deep neural networks.
\newblock In {\em Neural Information Processing Systems (NIPS)}, 2012.

\bibitem{YehLimChen18}
R.A. Yeh, T.~Y. Lim, C. Chen, A.~G. Schwing, M. Hasegawa-Johnson, and M.~N. Do.
\newblock Image restoration with deep generative models.
\newblock In {\em ICASSP}, 2018.

\bibitem{ZhaZuoCheMenZha17}
K. Zhang, W. Zuo, Y. Chen, D. Meng, and L. Zhang.
\newblock Beyond a gaussian denoiser: Residual learning of deep cnn for image
  denoising.
\newblock {\em IEEE Trans. Image Processing}, 26(7):3142 -- 3155, 2017.

\bibitem{ZorWei11}
D. Zoran and Y. Weiss.
\newblock From learning models of natural image patches to whole image
  restoration.
\newblock {\em International Conference on Computer Vision (ICCV)}, 2011.

\end{thebibliography}

\end{document}



\maketitle

\section{Proof of Lemma 1}

\begin{lemma}\label{lem1}
For any $\mathbf{N}$ with the assumptions in [Manuscript, Section 3.1] and $\Xhb(\Z)$ that has the form [Manuscript, Eq.(2)] with $d\in\{1,2\}$, \begin{eqnarray}
\E\Ellb_n(\Z,\Xhb(\Z);\sigma^2) = \E\Lb_n(\x,\Xhb(\Z)).\label{eq:unbiased}
\end{eqnarray}
Moreover, when $\mathbf{N}$ is white Gaussian, then, $\Ellb_n(\Z,\Xhb(\Z);\sigma^2)$ coincides with the SURE \cite{stein1981estimation}.
\end{lemma}
\emph{Proof:} We note the expectation of the $i$-th summand in [Manuscript, Eq.(3)] is 
\begin{align}
\textstyle &\frac{1}{n}\E\Big[(Z_i-\hat{X}_i(\Z))^2+\sigma^2(\sum_{m=1}^d2^ma_{m,i}Z_i^{m-1}-1)\Big]\nonumber\\
\textstyle =&\frac{1}{n}\E\Big[\E\big[(Z_i-\hat{X}_i(\Z))^2+\sigma^2(\sum_{m=1}^d2^ma_{m,i}Z_i^{m-1}-1)\big|\Z^{- i}\big]\Big]\label{eq:sup1}
\end{align}
Now, we divide into two cases, $d=1$ and $d=2$.

1) For $d=1$ (affine mapping), (\ref{eq:sup1}) becomes
\begin{align}
\textstyle =&\frac{1}{n}\E\Big[\E\big[(Z_i-(a_{1,i}Z_i+a_{0,i}))^2 \nonumber\\
&\quad\quad\quad\quad\quad\quad\quad\quad+\sigma^2(\sum_{m=1}^12^ma_{m,i}Z_i^{m-1}-1)\big|\Z^{- i}\big]\Big]\nonumber\\
\textstyle =&\frac{1}{n}\E\Big[\E\big[(Z_i^2-\sigma^2)+(a_{1,i}Z_i+a_{0,i})^2 \nonumber\\
&\quad\quad\quad\quad\quad\quad\quad\quad-2a_{1,i}(Z_i^2-\sigma^2)-2a_{0,i}Z_i|\Z^{- i}\big]\Big]\nonumber\\
\textstyle =&\frac{1}{n}\E\Big[\E\big[(x_i^2+(a_{1,i}Z_i+a_{0,i})^2-2a_{1,i}Z_ix_i-2a_{0,i}x_i|\Z^{- i}\big]\Big]\nonumber\\
\textstyle =&\frac{1}{n}\E\Big[\E\big[(x_i-(a_{1,i}Z_i+a_{0,i}))^2|\Z^{- i}\big]\Big]\nonumber\\
\textstyle =&\frac{1}{n}\E\Big[\E\big[(x_i-\hat{X}_i(\Z))^2|\Z^{-i}\big]\Big]\label{eq:lem_d1}\\
=&\frac{1}{n}\E\Big[x_i-\hat{X}_i(\Z)\Big]^2.\nonumber
\end{align}

2) For $d=2$ (polynomial mapping), (\ref{eq:sup1}) becomes
\begin{align}
\textstyle =&\frac{1}{n}\E\Big[\E\big[(Z_i-(a_{2,i}Z_i^2+a_{1,i}Z_i+a_{0,i}))^2 \nonumber\\
&\quad\quad+\sigma^2(\sum_{m=1}^22^ma_{m,i}Z_i^{m-1}-1)\big|\Z^{- i}\big]\Big]\nonumber\\
\textstyle =&\frac{1}{n}\E\Big[\E\big[(Z_i^2-\sigma^2)+(a_{2,i}Z_i^2+a_{1,i}Z_i+a_{0,i})^2 \nonumber\\
&\quad\quad-2a_{2,i}(Z_i^3-2Z_i\sigma^2)-2a_{1,i}(Z_i^2-\sigma^2)-2a_{0,i}Z_i|\Z^{- i}\big]\Big]\nonumber\\
\textstyle =&\frac{1}{n}\E\Big[\E\big[(x_i^2+(a_{2,i}Z_i^2+a_{1,i}Z_i+a_{0,i})^2 \nonumber\\
&\quad\quad-2a_{2,i}Z_i^2x_i-2a_{1,i}Z_ix_i-2a_{0,i}x_i|\Z^{- i}\big]\Big]\nonumber\\
\textstyle =&\frac{1}{n}\E\Big[\E\big[(x_i-(a_{2,i}Z_i^2+a_{1,i}Z_i+a_{0,i}))^2|\Z^{- i}\big]\Big]\nonumber\\
\textstyle =&\frac{1}{n}\E\Big[\E\big[(x_i-\hat{X}_i(\Z))^2|\Z^{-i}\big]\Big]\label{eq:lem_d2}\\
=&\frac{1}{n}\E\Big[x_i-\hat{X}_i(\Z)\Big]^2.\nonumber
\end{align}
Note (\ref{eq:lem_d1}) and (\ref{eq:lem_d2}) are from the specific form of $\hat{X}_i(\Z)$, the fact $\{a_{m,i}\}$'s are independent of $Z_i$ given $\Z^{-i}$, and 
\begin{eqnarray}
\E(Z_i^3-2Z_i\sigma^2|\Z^{- i})&=&\E(x_iZ_i^2|\Z^{-i})\nonumber\\
\E(Z_i^2-\sigma^2|\Z^{- i})&=&\E(x_iZ_i|\Z^{-i})\nonumber\\
&=&\E(x_i^2|\Z^{-i})\nonumber\\
\E(Z_i|\Z^{-i})&=&\E(x_i|\Z^{-i}),\nonumber
\end{eqnarray}
which hold due to the assumptions on $\mathbf{N}$ in Section 3.1 and $\x$ being an individual image. Thus, we obtain the Lemma by obtaining the unbiasedness for all $i=1,\ldots,n$.

Furthermore, when $\mathbf{N}$ is i.i.d. Gaussian, then the SURE of $\Lb_n(\x,\Xhb(\Z))$ becomes
\begin{eqnarray}
-\sigma^2+\frac{1}{n}\|\Z-\Xhb(\Z)\|_2^2+\frac{2\sigma^2}{n}\sum_{i=1}^n\frac{\partial \hat{X}_i(\Z)}{\partial Z_i},
\end{eqnarray}
which is equivalent to [Manuscript, Eq.(3)] when $\hat{X}_i(\Z)=\sum_{m=0}^da_m(\Z^{-i})Z_i^m$ with $d\in\{1,2\}$. \ \qed

\begin{table*}[h]\caption{PSNR(dB) on \emph{Image13} and \emph{BSD68}.}
\centering
\smallskip\noindent
\begin{tabular}{|c||c||c|c|c|}
\hline
Data$\backslash$Alg.     & \fcaidesft   & \fcaidesfta & \fcaidesd & \fcaidesftd \\ \hline \hline
\emph{Image13} & \textbf{29.33}    & 20.46  & 26.99 & 27.69 \\ \hline
\emph{BSD68 (avg.)} & \textbf{29.31}  & 19.20  & 27.66 & 27.68 \\ \hline
\end{tabular}
    \label{table:results}
\end{table*}
\vspace{-.12in}

\section{The unbiasedness of $\Ellb_n(\cdot)$}

\begin{figure}[h]
\vspace{-.15in}
    \centering
    \includegraphics[width=0.96\linewidth]{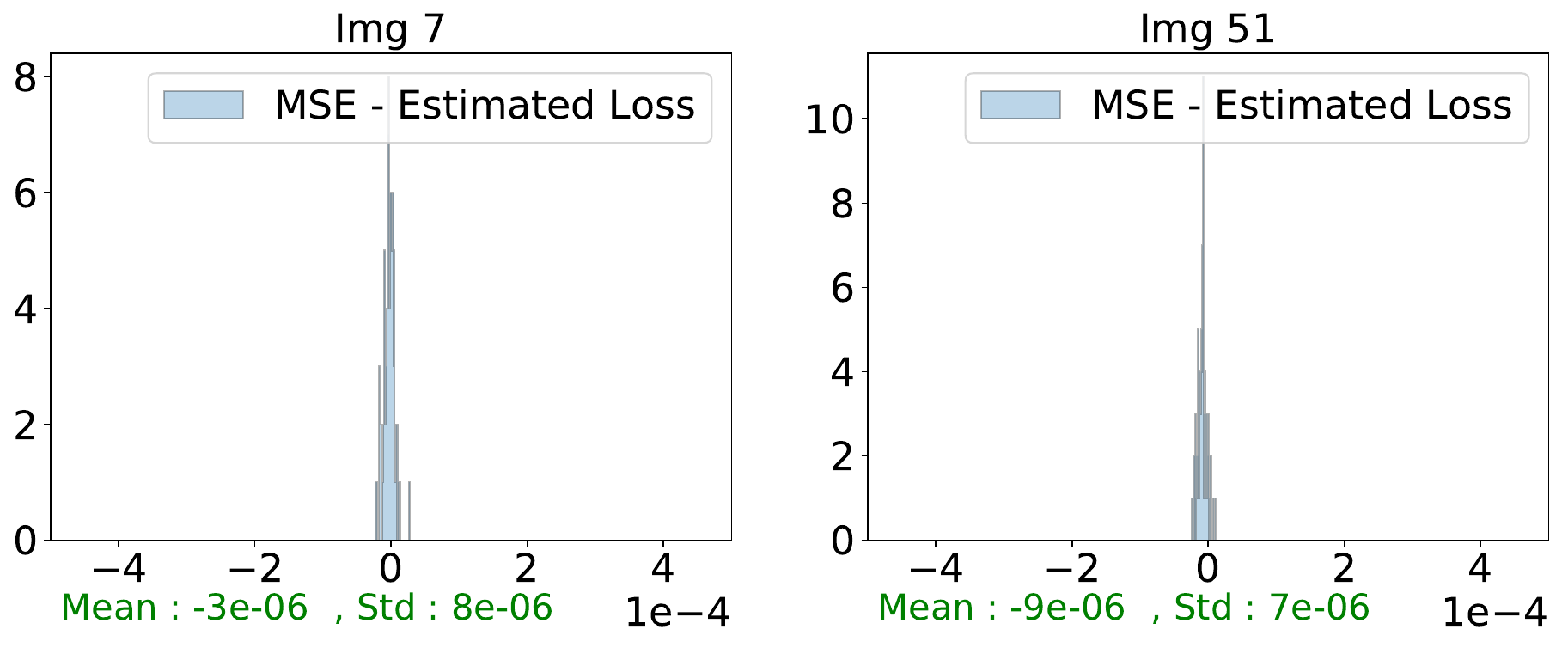}
    \caption{The difference between MSE and Eq.(3) for \fcaideb. }
    \label{fig:unbiased}\vspace{-.1in}
\end{figure}

Here, we also experimentally verify the unbiasedness of $\Ellb_n(\cdot)$ that is analytically shown in Lemma \ref{lem1}. Figure \ref{fig:unbiased} shows the histograms of differences between MSE and [Manuscript, Eq.(3)] of the \fcaideb model, for 100 independent noise realizations ($\sigma=25$) on two randomly selected images in BSD68. The mean of the difference clearly concentrates on 0 (\ie, unbiased), and the standard deviation is also extremely small, for both images. 

\section{Supplementary for Section 4.2}

\subsection{Ablation study for data augmentation}

\begin{figure}[ht]
    \centering
    \includegraphics[width=0.8\linewidth]{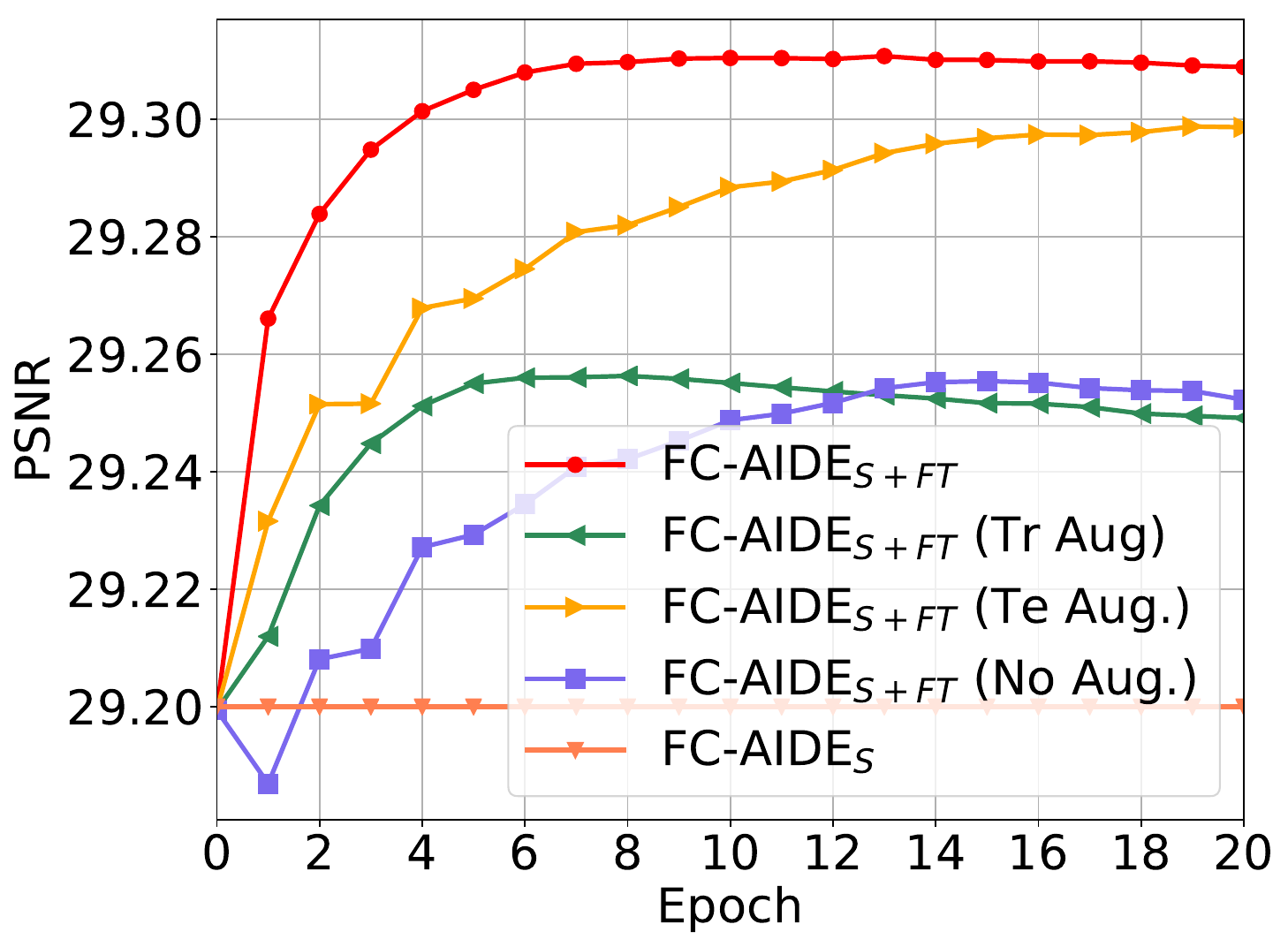}
    \caption{Ablation study on augmentations on BSD68 ($\sigma=25$).}
    \label{fig:augmentations}\vspace{-.1in}
\end{figure}

Figure \ref{fig:augmentations} shows two additional results that use the \emph{self-ensemble} data augmentation only for the training (Tr Aug) or testing (Te Aug) of the fine-tuning. Note ``Te Aug'' leads to more improvement, while the full augmentation, which is employed by our \fcaidesft, leads to the highest PSNR and stable convergence.

From Figure \ref{fig:augmentations}, we observe that the maximum PSNR of \fcaidesft is achieved around epoch 5, but even with a \emph{single} epoch, the PSNR significantly improves over \fcaides. Each epoch takes about 3 seconds, and early stopping can lead to the accuracy-complexity tradeoff. Moreover, the running time of each epoch for ``No Aug'' was 1.8 second, hence, the running time of the partial augmentation schemes
lie in between 1.8s and 3s. 

\subsection{Hyperparameter selection for $\ell_2$-SP}

\begin{figure}[h]
\centering
\includegraphics[width=.4\textwidth]{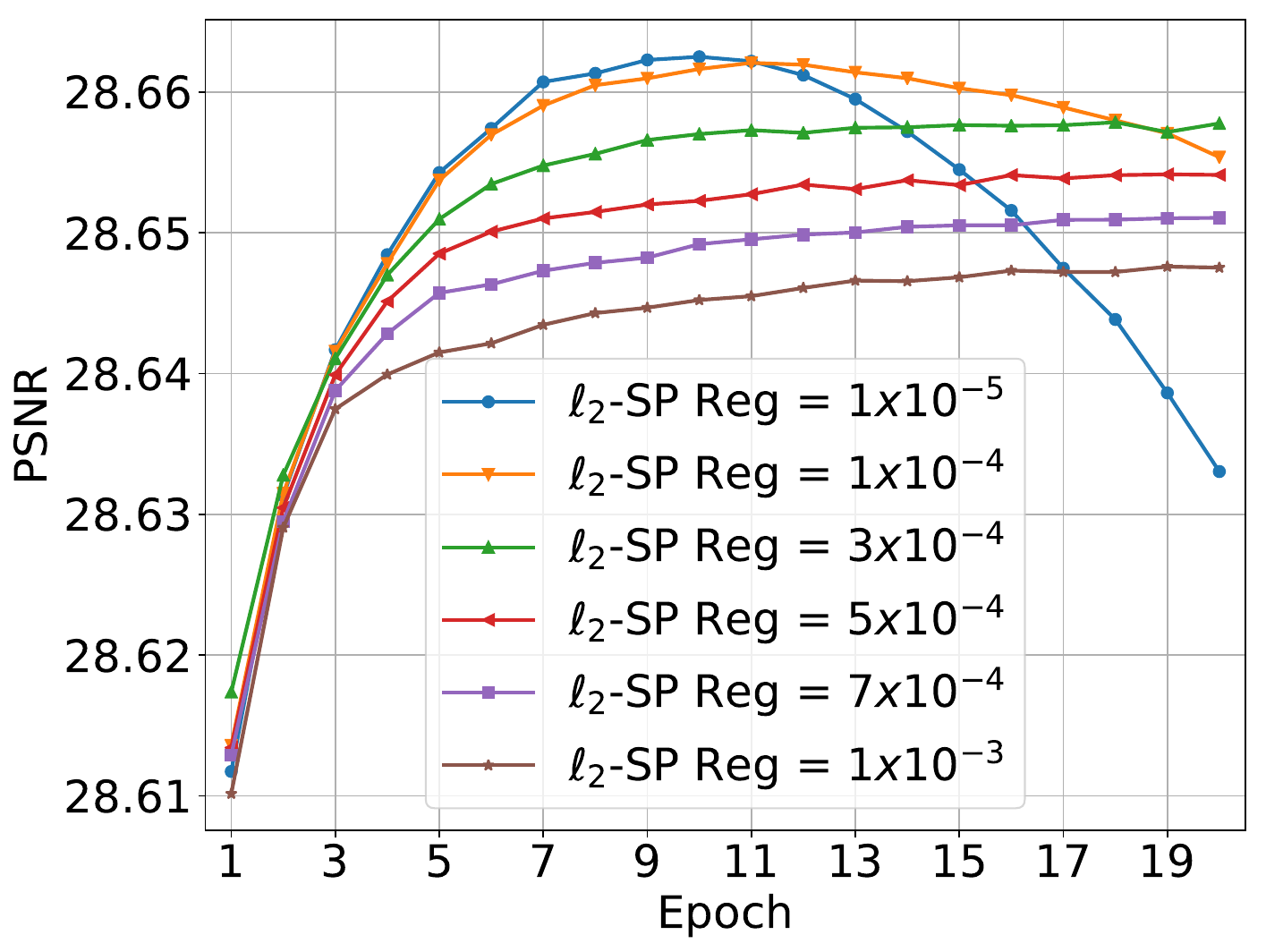}
\caption{Fine-tuning result on the validation set ($\sigma=25$)}\label{fig:val32}
\end{figure}

\begin{table}[h]
\centering
\caption{Selected regularization strength for $\ell_2$-SP and the stopping epoch for adaptive fine-tuning for each $\sigma$.}\vspace{.1in}
\label{table:val32_result}
\begin{tabular}{|c||c|c|}
\hline
              & $\lambda$ for $\ell_2$-SP  & Stopping epoch \\ \hline
              \hline
$\sigma = 15$ & $1\times10^{-4}$             & $5$                    \\ \hline
$\sigma = 25$ & $3\times10^{-4}$             & $4$                    \\ \hline
$\sigma = 30$ & $5\times10^{-4}$             & $3$                    \\ \hline
$\sigma = 50 $& $2\times10^{-3}$             & $2$                    \\ \hline
$\sigma = 75 $& $5\times10^{-3}$             & $1$                    \\ \hline
\end{tabular}
\end{table}


As mentioned in [Manuscript, Section 5.1], we used a separate validation set that consists of 32 natural images from BSD300 \cite{martin2001database} for selecting the hyper-parameters in our fine-tuning step (\ie, the stopping epoch and the regularization parameter for $\ell_2$-SP). Note the validation images do not overlap with our training and test images. We carried out the validation for each noise level $\sigma=\left\{15,25,30,50,75\right\}$ separately and selected the best hyper-parameters that gave the best trade-off between the PSNR and the robustness of the curve. Figure \ref{fig:val32} shows the results for $\sigma=25$, for example. Note the PSNR results are not very different among the hyper-parameter choices, and the selection results for all noise levels are given in Table \ref{table:val32_result}. These hyper-parameters were used for all our experiments in the paper.

\section{Supplementary for Section 5.3}

\begin{figure}[h]
\centering
\includegraphics[width=.4\textwidth]{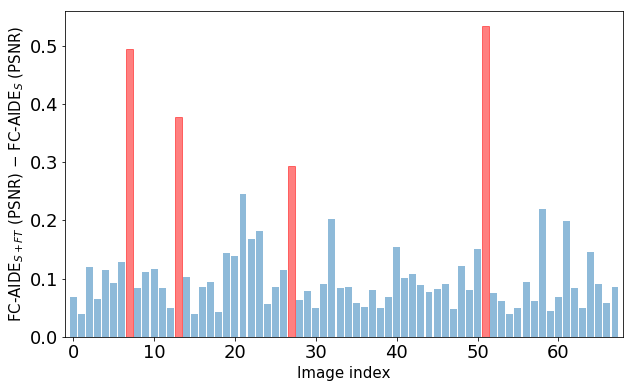}
\caption{Improvement on BSD68}\label{fig:bsd68_improvements}
\end{figure}

Figure \ref{fig:bsd68_improvements} shows the PSNR differences between \fcaidesft and \fcaides for each test image in BSD68 with $\sigma=25$. Note the adaptive fine-tuning gives positive PSNR gains for \emph{all} the images, and the four red bars indicate the images with the most PSNR improvements that are visualized in [Manuscript, Figure 4].


\section{Supplementary for Section 5.4}

Here, we emphasize the importance of the polynomial coefficients of \fcaidesft for denoising. In Table \ref{table:results}, we report the PSNRs on \emph{Image13} (of BSD68) as well as the average PSNRs on the entire BSD68. Note the visualizations of the pixelwise coefficients are given in [Manuscript, Figure 7]. In the table, we compare \fcaidesft with several other baseline models; \fcaidesfta is a scheme that denoises only with the $a_0$ terms after learning \fcaidesft, \fcaidesd is a supervised-trained model with setting $a_1=a_2=0$, and \fcaidesftd is the model obtained by fine-tuning \fcaidesd using $\Ellb_n(\cdot)$ (Manuscript, Eq.(3)).

\begin{figure}[h]
    \centering
        \includegraphics[width=0.9\linewidth]{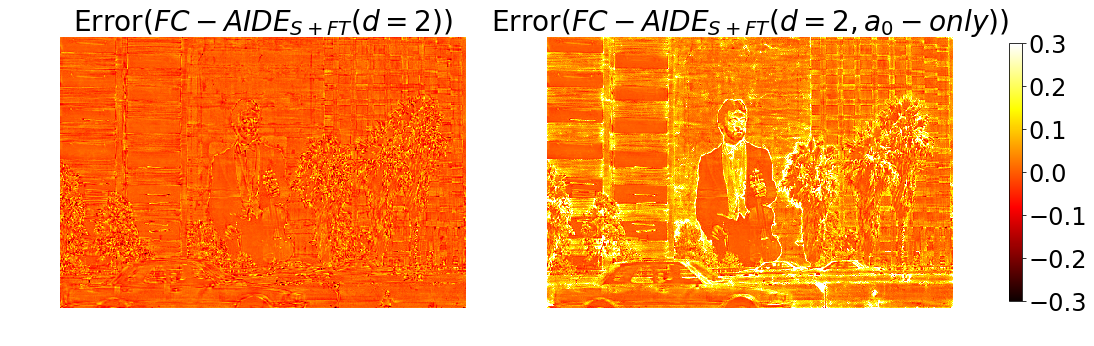}
    ~ 
    \caption{Pixelwise errors of \fcaidesft and \fcaidesfta. }\vspace{-.13in}
     \label{fig:error}
\end{figure}

Note \fcaidesfta and \fcaidesftd are different schemes, and they are \underline{\emph{not}} equivalent to the regular end-to-end scheme, since they both do not use $Z_i$ and are adaptively fine-tuned. From the table, we note that \fcaidesfta hardly does any denoising (as the PSNR of the noisy \emph{Image13} is 20.16dB), 
and \fcaidesftd is also much worse than \texttt{FC-AIDE}$_{\texttt{S+FT}}$. Figure \ref{fig:error} shows the pixelwise errors on \emph{Image13}, further demonstrating the importance of $a_1$ and $a_2$ in our polynomial model.

\section{Visualization}

\begin{figure*}[t]
\centering
\includegraphics[width=0.9\textwidth]{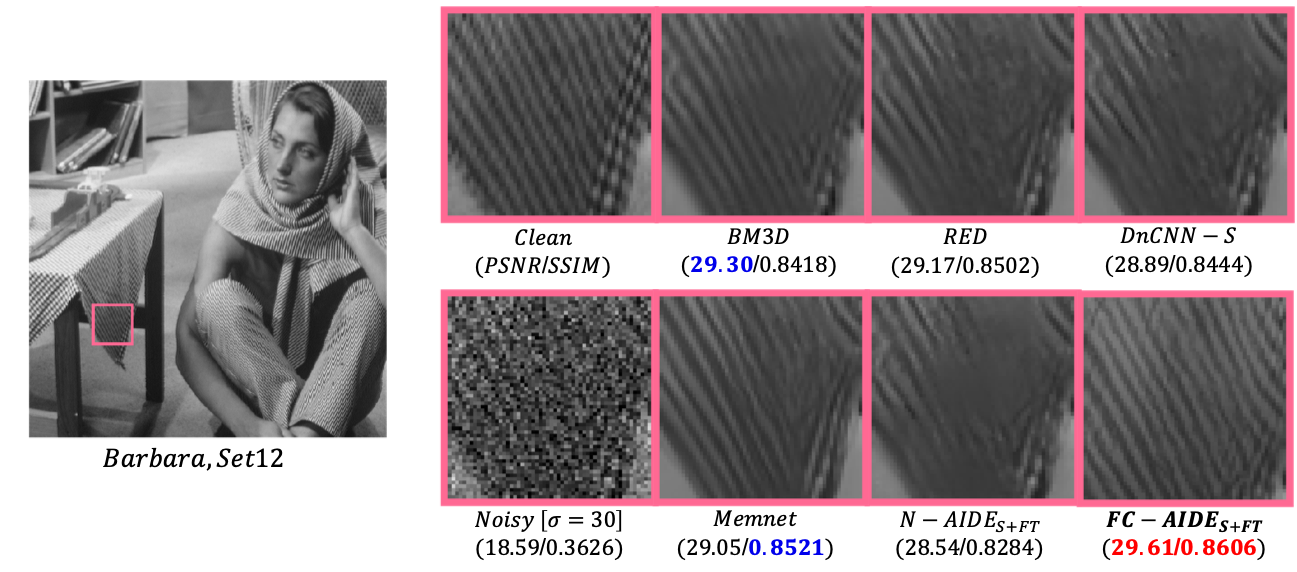}
\includegraphics[width=0.9\textwidth]{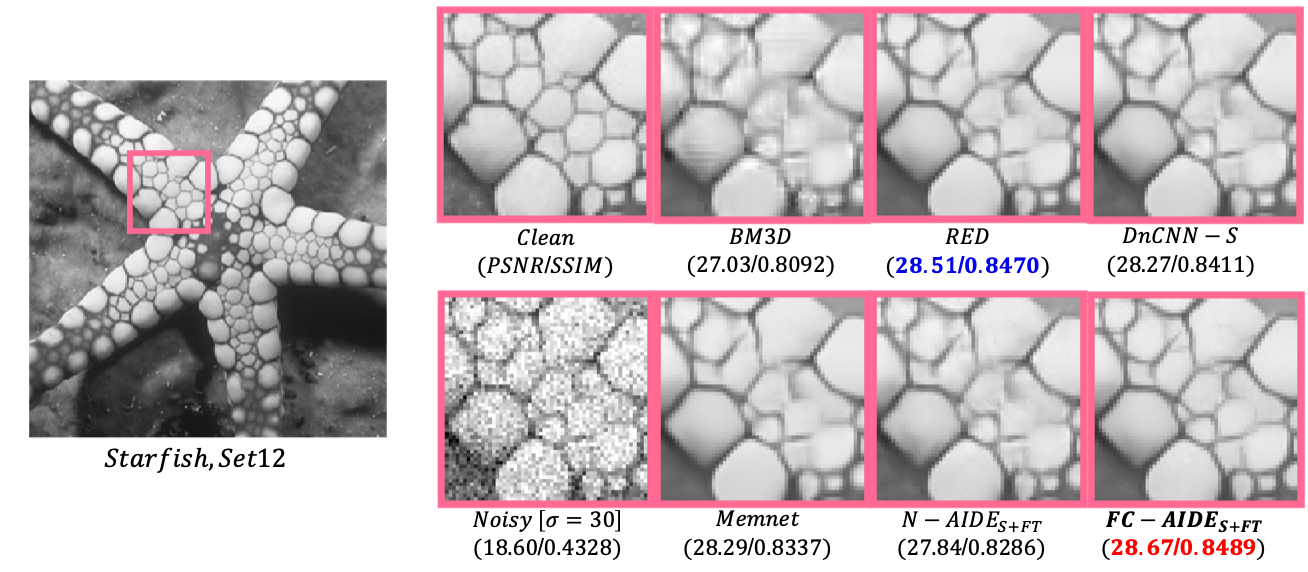}
\includegraphics[width=0.9\textwidth]{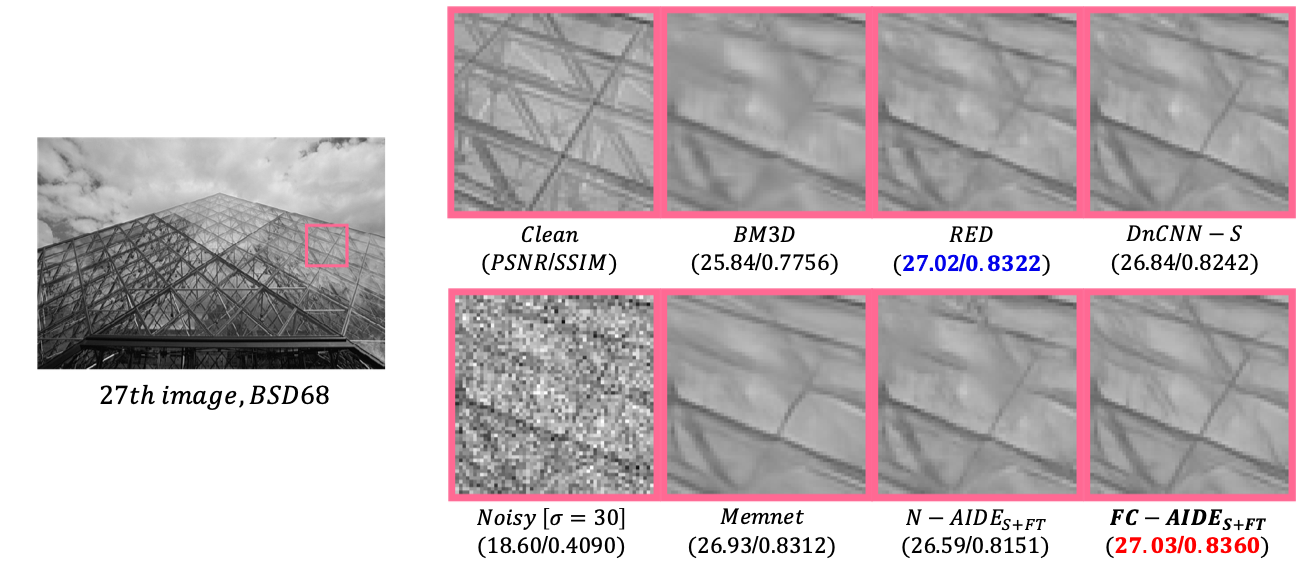}
\caption{Denoising results on Set12 and BSD68}\label{fig:set12_bsd68}
\end{figure*}

Figure \ref{fig:set5_clean} and \ref{fig:set12_clean}  show the clean images used for Set5 and Set12.
Moreover, in Figures \ref{fig:set12_bsd68}$\sim$\ref{fig:medical_gaussian}, we visualized the denoising results on sample images from our evaluation datasets, \ie, Set12, BSD68, Urban100, Manga109, \emph{BSD68/Laplacian} and \emph{Medical/Gaussian}. We compare our \fcaidesft with the most competitive state-of-the-art baselines and show the superiority of \fcaidesft both quantitatively and qualitatively.


\begin{figure*}[t]
\centering
\includegraphics[width=0.9\textwidth]{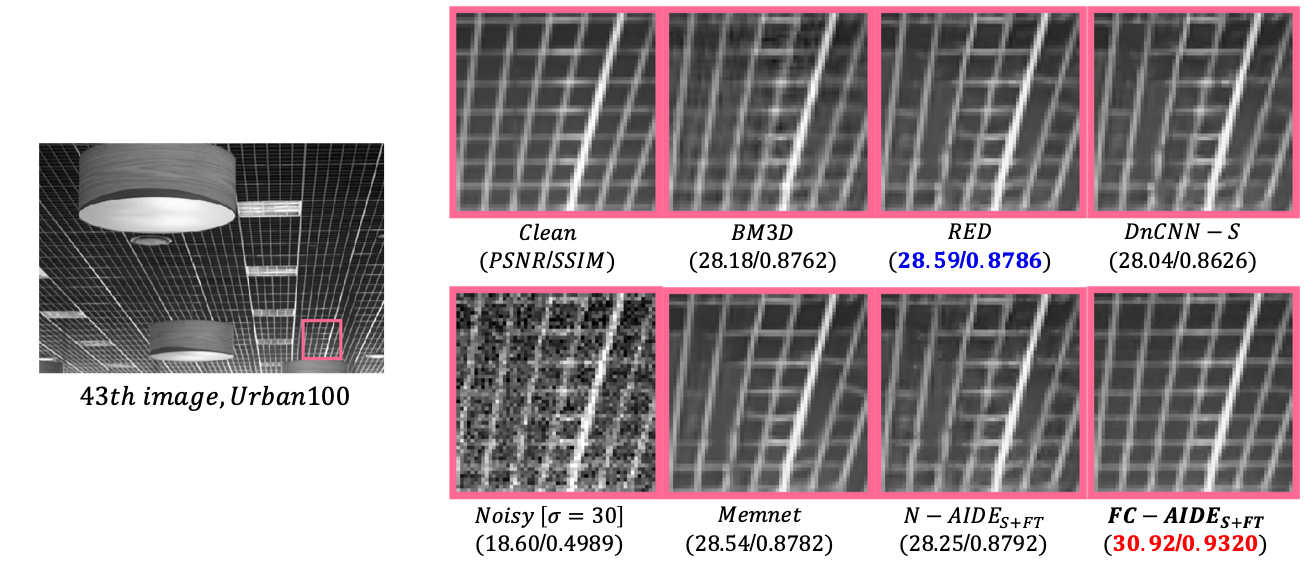}
\includegraphics[width=0.9\textwidth]{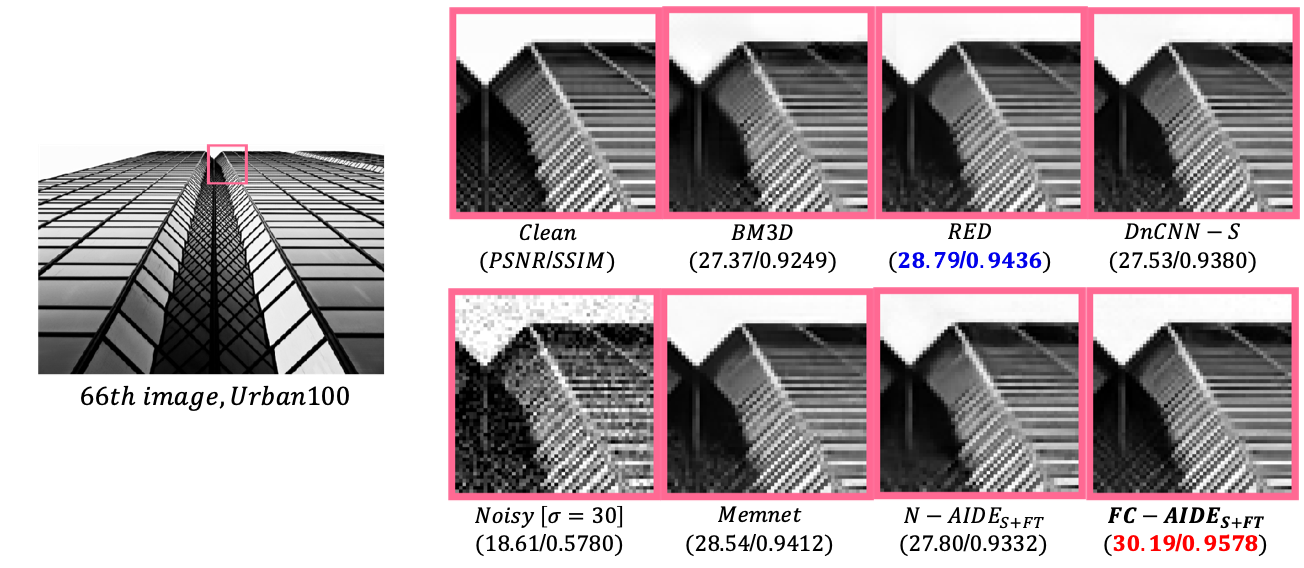}
\includegraphics[width=0.9\textwidth]{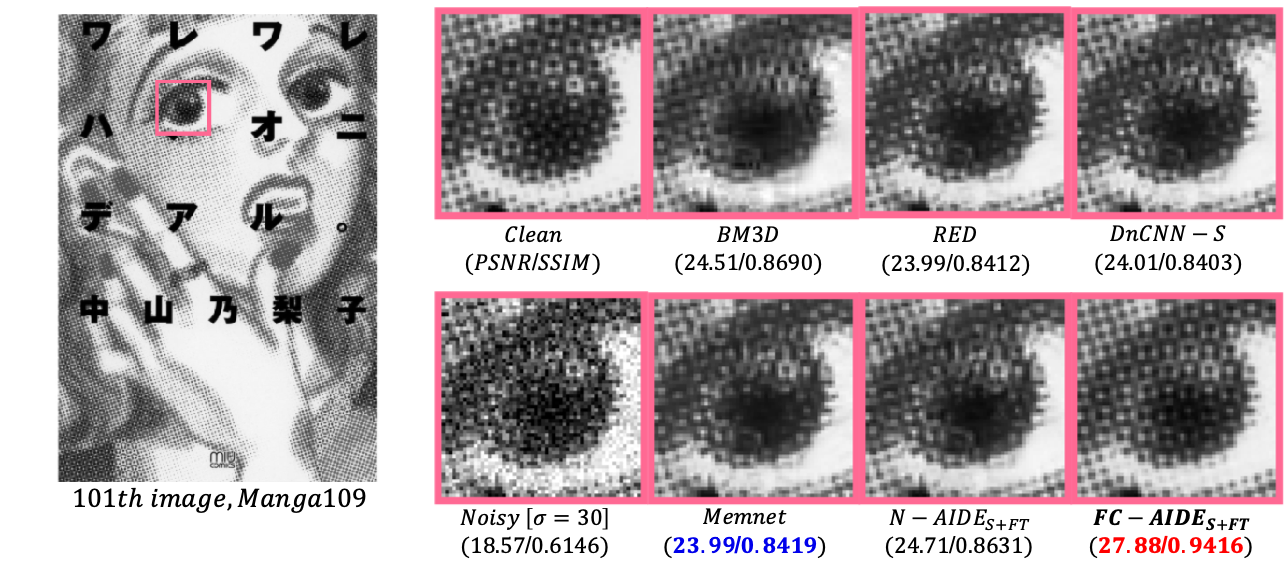}
\caption{Denoising results on Urban100 and Manga109}\label{fig:urban100_manga109}
\end{figure*}

\begin{figure*}[t]
\centering
\includegraphics[width=0.9\textwidth]{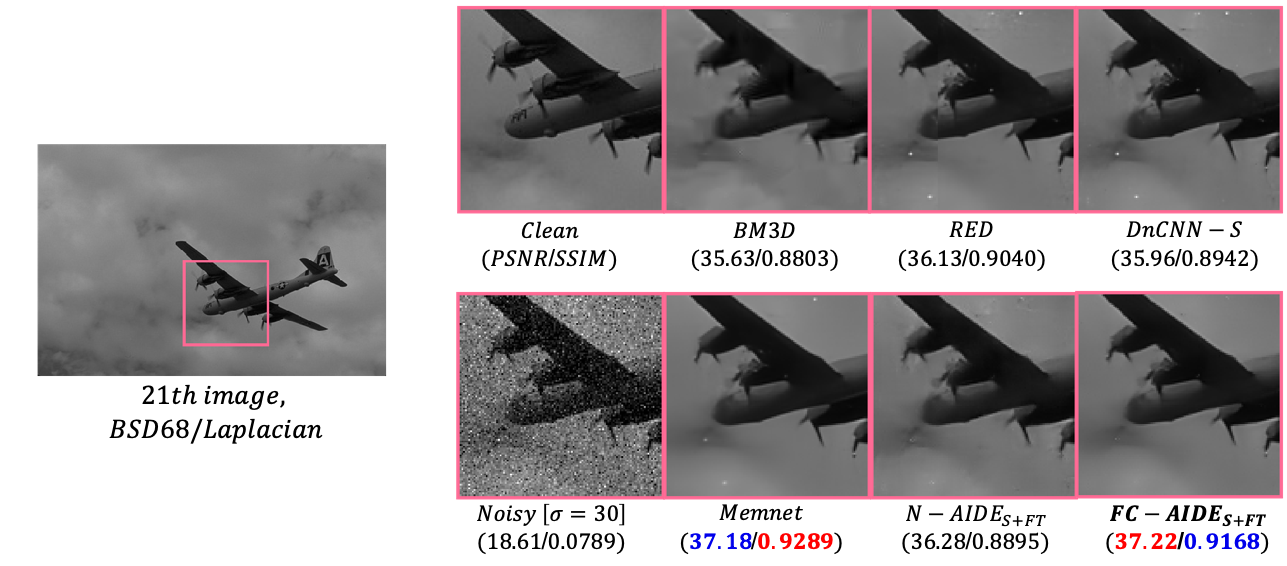}
\includegraphics[width=0.9\textwidth]{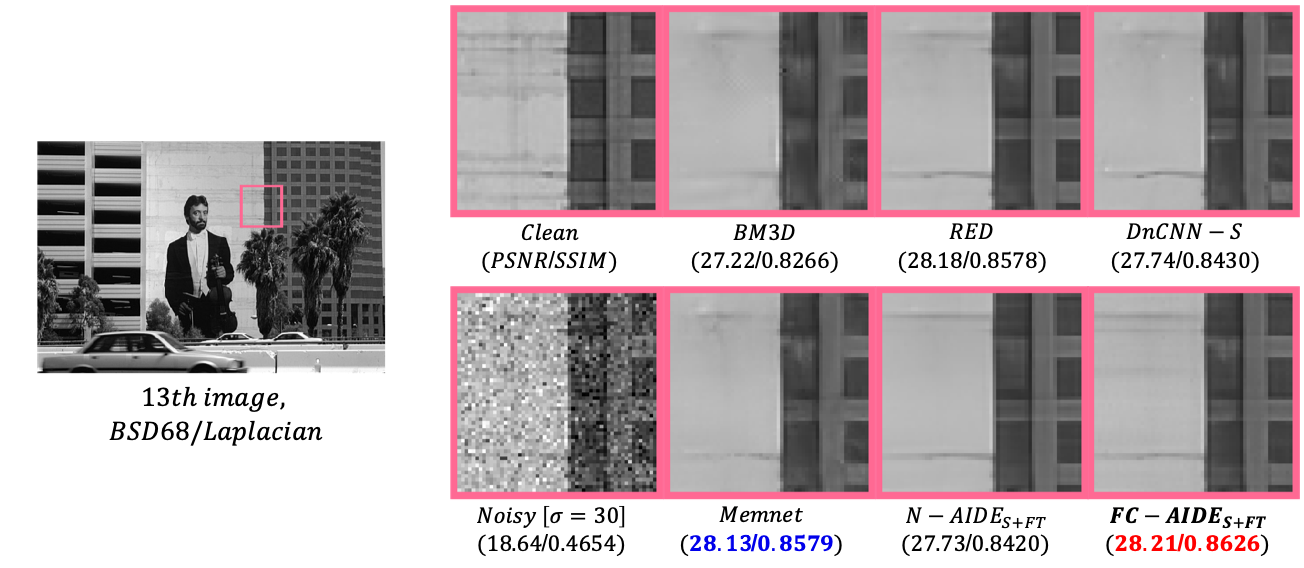}
\caption{Denoising results on \emph{BSD68/Laplacian}}\label{fig:bsd68_laplacian}
\end{figure*}

\begin{figure*}[t]
\centering
\includegraphics[width=0.9\textwidth]{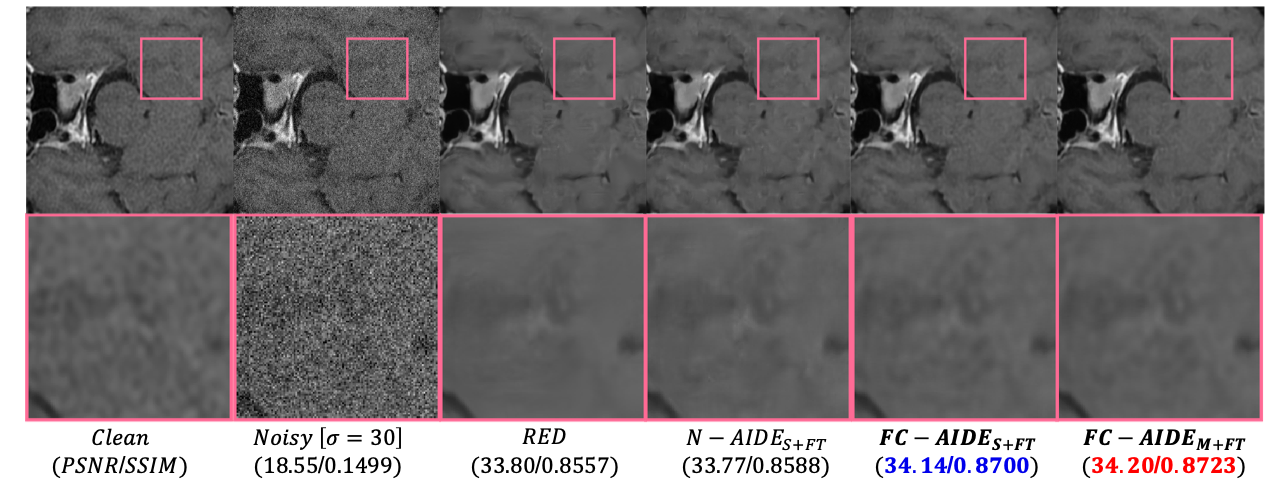}
\caption{Denoising results on \emph{Medical/Gaussian}}\label{fig:medical_gaussian}
\end{figure*}

\begin{figure*}[t]
\centering
\includegraphics[width=.6\textwidth]{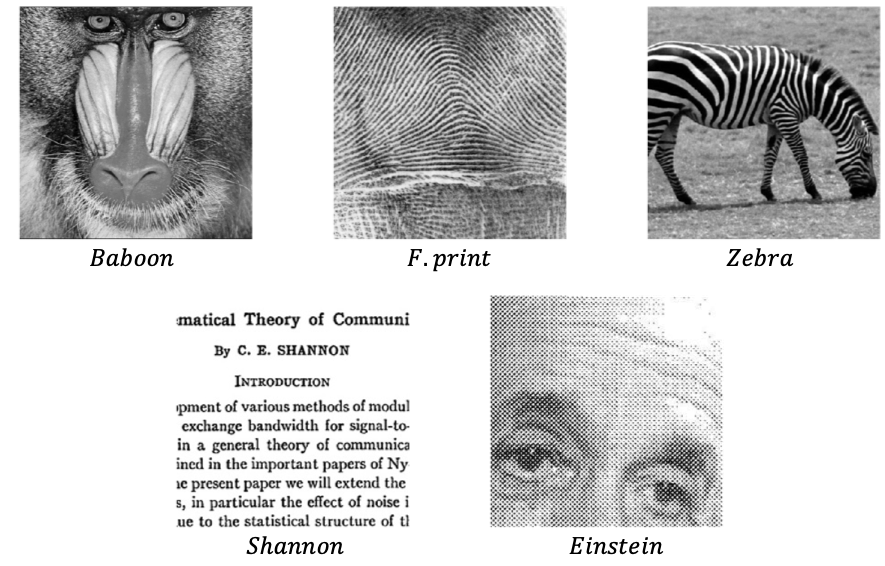}
\caption{Visualization on Set5.}\label{fig:set5_clean}
\end{figure*}

\begin{figure*}[t]
\centering
\includegraphics[width=\textwidth]{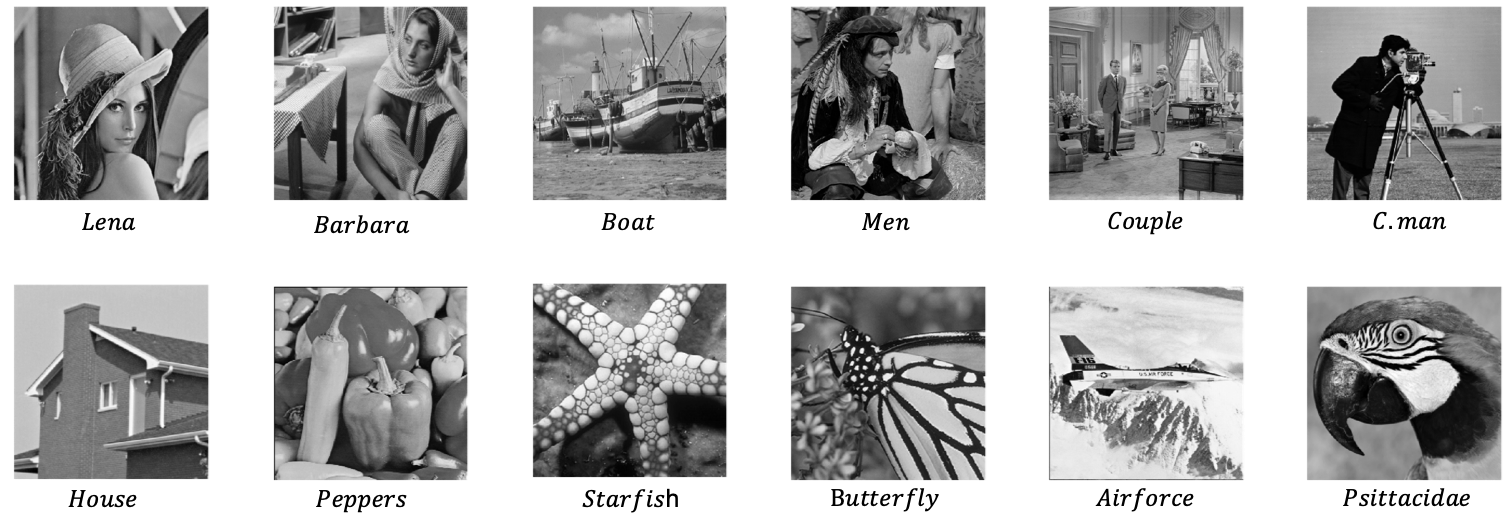}
\caption{Visualization on Set12.}\label{fig:set12_clean}
\end{figure*}

\newpage


\bibliographystyle{ieee_fullname}
\bibliography{egbib}